%% file: main.tex
\documentclass[11pt, a4paper, logo, copyright, nonumbering]{xiaomi}

\usepackage[authoryear, sort&compress, round]{natbib}
\usepackage{dblfloatfix}
\usepackage{dcolumn,placeins}
\usepackage{ulem}
\usepackage{caption}
\usepackage{dramatist}
\usepackage{xspace}
\usepackage{pifont} 
\usepackage{multirow}
\usepackage{tcolorbox}
\usepackage{xltabular}
\usepackage{longtable}
\usepackage{hyperref}
\usepackage{wrapfig}

\usepackage{amsfonts}
\usepackage{amsmath}
\usepackage{amssymb}
\usepackage{lineno}
\usepackage{multirow}
\usepackage{adjustbox}
\definecolor{xiaomiorange}{HTML}{FF6901}
\usepackage[bottom]{footmisc}
\usepackage{dsfont}

\usepackage{setspace}
\usepackage{makecell}
\usepackage{graphicx}
\usepackage{subcaption}
\usepackage{multicol} 

\newcommand{\mimox}{MiMo-V2.6}
\newcommand{\pro}{\mimox{}-Pro}
\newcommand{\flash}{\mimox{}-Flash}

\definecolor{xiaomiorange}{HTML}{FF6901}

\title{\centering \mimox{}: Scaling Reinforcement Learning Towards Self-Improvement}

\author{
 LLM-Core Xiaomi
}

\begin{abstract}
Reinforcement learning (RL) is the central training paradigm for advancing large foundation models towards self-improvement. This report introduces the MiMo-V2.6 series, an omni-modal family that pushes the frontier of model intelligence by scaling RL compute. Prior to RL, we conduct mid-training on a broad multimodal corpus to provide ample exploration space, and build a solid infrastructure on the pretrained hybrid-SWA architecture to support subsequent scale-up. We scale RL compute along three dimensions: (1) larger batches and higher throughput, with an asynchronous training that consumes 1,568 samples and 2.7$\sim$3.7B tokens per step at context lengths of up to 1M; (2) more diverse and complex environments, spanning code, general, visual, and cyber domains under a mixture of agent harnesses; and (3) more grader compute, via groupwise agentic grading that yields more accurate reward signals for long-horizon tasks and steers the model towards shorter, more token-efficient solutions. To keep training stable at scale, we freeze the MoE router and establish a multi-layer defense against reward hacking. We further build infrastructure for mixed-task agentic RL, including a unified trajectory representation, high-concurrency multi-framework rollout, decoupled control and data planes, and training–inference consistency. We open-source the training dynamics, RL environments, and RL framework to facilitate reproduction and further research on scaled RL and model self-improvement.

\end{abstract}

\begin{document}

\begingroup
\patchcmd{\maketitle}{\vskip20pt}{\vskip10pt}{}{}
\patchcmd{\maketitle}{\vskip30pt}{\vskip12pt}{}{}
\patchcmd{\abscontent}{\vspace{5ex}}{\vspace{2ex}}{}{}
\maketitle
\endgroup

\par\vspace{0.5em}
\noindent
\begin{minipage}{\linewidth}
    \centering
    \captionsetup{skip=4pt, font=small}
    \includegraphics[width=0.96\linewidth]{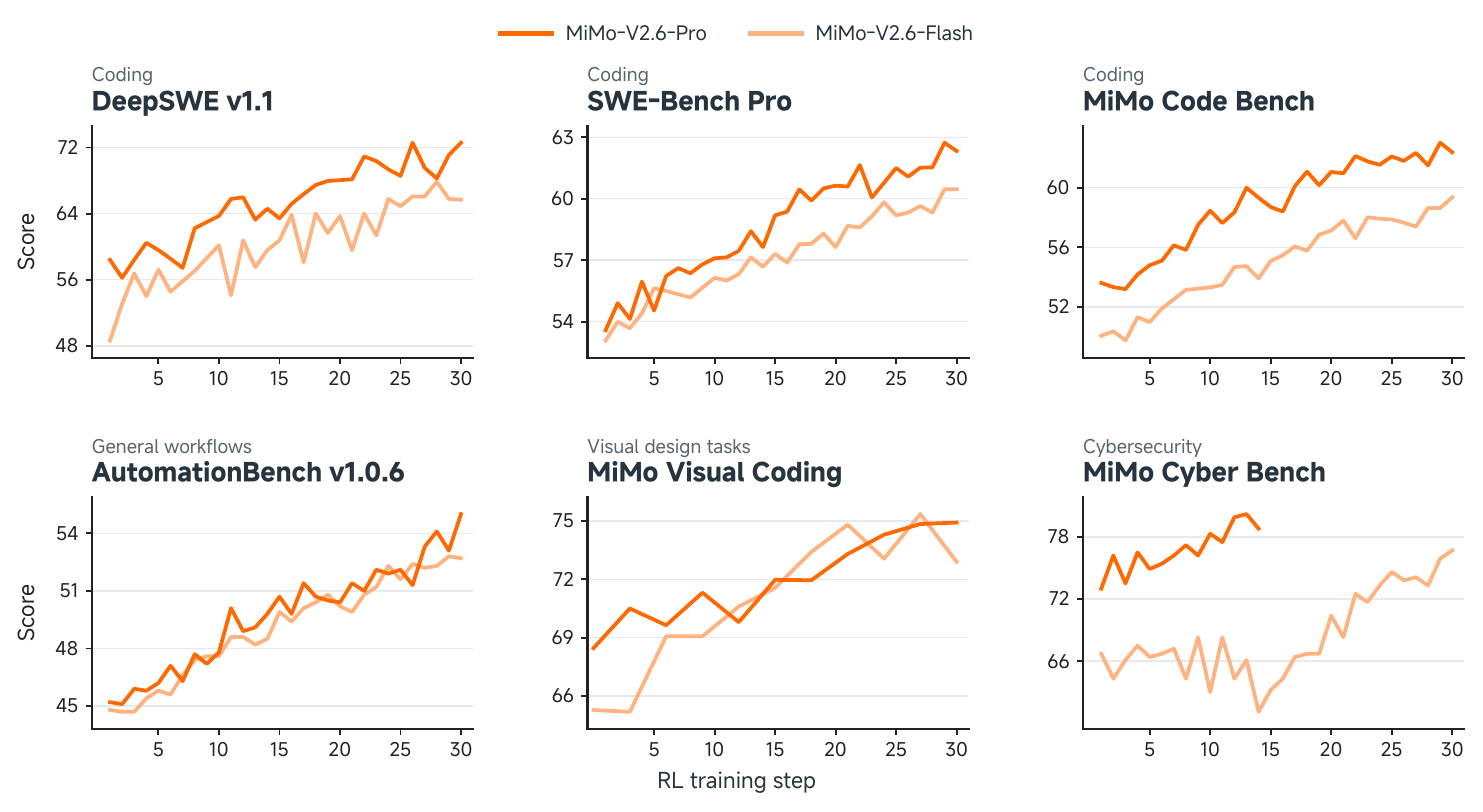}
    \captionof{figure}{Benchmark score per task of MiMo-V2.6-Pro and MiMo-V2.6-Flash throughout RL training.}
    \label{fig:rl_teaser}
\end{minipage}
\par

\newpage

\begin{spacing}{0.85}
\tableofcontents
\end{spacing}

\newpage

\input{1-intro}

\input{2-arch}

\input{3-pretrain}

\input{4-rlscale}

\input{5-yoro}

\input{6-mopd}

\input{7-infra}

\input{8-open}

\section{Conclusion}
This report presents the MiMo-V2.6 series and a practical approach to advancing foundation models through large-scale agentic reinforcement learning. Building on an omni-capable foundation and agent-centric mid-training, we scale RL along three dimensions: training batch size and throughput, the diversity and complexity of environments and agent harnesses, and the compute devoted to groupwise agentic grading. These advances are supported by asynchronous training, mixed-task rollout infrastructure, training--inference consistency mechanisms, and safeguards against training drift and reward hacking. Together, they enable sustained optimization over long-horizon interactions while improving solution quality and token efficiency. Evaluations across public and internal benchmarks demonstrate the effectiveness of this approach, with MiMo-V2.6 achieving competitive performance against frontier models across a broad range of agentic tasks.

To make this direction accessible to the research community, we release MiMo-V2.6-Distill-Qwen-9B alongside curated task environments, verifiers, an end-to-end RL framework, and composable mini-harnesses. Consistent RL gains across domains and harnesses highlight the value of these resources as a shared foundation for further experimentation. Our work takes a practical step toward model self-improvement, emphasizing the joint importance of broad exploration, informative feedback, and scalable training systems. We hope these open resources support reproducible research and cumulative progress toward increasingly capable, general-purpose agents.

\bibliography{reference}


\appendix

\newpage

\input{author}

\end{document}

%% file: 1-intro.tex
\section{Introduction}
Recursive self-improvement (RSI) envisions models that expand their capabilities through sustained exploration and feedback.
Realizing this vision requires agents, which couple models with interactive environments and thereby supply the multi-step trajectories and feedback signals that self-improvement depends on.
Towards this goal, scaling reinforcement learning (RL) on complex agentic tasks thus opens a concrete path.
However, realizing this potential faces two major challenges. First, RL for foundation models requires suitable model architectures and a sufficiently rich exploration space for agents. Second, scaling RL requires sophisticated solutions for infrastructure, environments and grader.
In this report, we introduce the \mimox{} series, including \pro{}, a 1.02T-parameter Mixture-of-Experts model with 42B active parameters, and \flash{}, a 310B-parameter Mixture-of-Experts model with 15B active parameters, to bridge these gaps, taking a practical step in large-scale RL.

The architecture, pre-training, and mid-training of \mimox{} jointly establish a powerful and efficient foundation model. 
Motivated by the need to preserve global context at low computational cost, \mimox{} builds a hybrid sparse MoE Transformer backbone that interleaves Local Sliding Window Attention (SWA) with Global Attention (GA), augmented by a lightweight visual encoder, audio encoders.
In addition, large-scale pre-training equips the model with extensive knowledge and versatile multimodal understanding across text, audio, and video. 
We further introduce an agent-centric mid-training phase that expands the exploration space for agentic tasks, enabling the model to discover more effective task-solving trajectories during post-training.

Next, we scale the RL compute through a systematic, co-designed framework spanning three key components.
First, we scale batch size and training throughput through fully asynchronous training, processing thousands of long-horizon rollouts and billions of tokens per step at context lengths of up to 1M.  
Second, we scale the diversity and complexity of RL environments across coding, general, visual, and cybersecurity tasks, using diverse agent harnesses and strengthening safeguards against reward hacking.
Varying both the harness and the task improves the model generalization. 
Third, we introduce groupwise agentic grading to provide more informative reward signals beyond binary test cases. Instead of assigning the same reward to all solutions that pass the test cases, we compare solutions within each group to distinguish problem-solving quality and behaviors. Through our proposed Groupwise Reward Synthesis (GRS) and Groupwise Advantage Redistribution (GAR), these fine-grained distinctions are converted into more informative learning signals, guiding the model toward more accurate and efficient solutions.

Realizing these at scale presents both research and engineering challenges for infrastructure.
To support flexible agentic interaction scenarios, we design a unified trajectory representation and a penalty mechanism that together refine the learning signal.
To handle large training batches, we sustain high-concurrency interaction across diverse agent frameworks through a pool of harness, and decouple the control plane from the data plane to buffer and transfer massive trajectories carrying routing and multi-modal payloads.
A sample mixing mechanism, working together with dynamic sampling and partial rollout, stabilizes per-task sample composition in train batches.
We align the training and inference engines on MoE routing and top-\textit{p} sampling candidate sets, and optimize both engines for RL workloads.

Scaling RL compute substantially unlocks model potential on both verifiable tasks such as coding and less verifiable tasks such as web development.
As shown in Figure~\ref{fig:rl_teaser}, both MiMo-V2.6-Pro and MiMo-V2.6-Flash steadily improve their performance across the reported benchmarks as training steps increase.
Moreover, these gains are not confined to a single domain, but hold consistently across diverse tasks spanning coding on DeepSWE~\citep{DBLP:journals/corr/abs-2607-07946}, general workflows on AutomationBench~\citep{shepard2026automationbench}, visual tasks on MiMo Visual Coding, and cybersecurity on MiMo Cyber Bench.
These gains indicate that \mimox{} is capable of stronger complex problem solving and more reliable performance in daily co-work scenarios, reflecting more general agentic intelligence.

To promote research in agentic RL, we open-source a comprehensive suite covering the key components of the RL development stack, including the lightweight model MiMo-V2.6-Distill-Qwen-9B, curated task environments with verifiers across multiple domains, an end-to-end RL training framework, and a composable mini-harness for flexible agentic interactions. Together, these components provide a unified and accessible platform for studying agentic RL under diverse tasks, environments, and agent configurations, while reducing the engineering barriers to reproducing and extending large-scale RL experiments. Experiments with MiMo-V2.6-Distill-Qwen-9B demonstrate consistent RL gains across diverse tasks and agent harnesses, validating the effectiveness and generality of the proposed suite. We hope this open-source suite can serve as a fair, strong, and reproducible baseline for the community, enabling systematic investigation of agentic RL and advancing research toward recursive self-improvement.

%% file: 2-arch.tex
\section{Architecture}

\subsection{Overall Architecture}

\begin{figure}[t!]
  \centering
  \includegraphics[width=\linewidth]{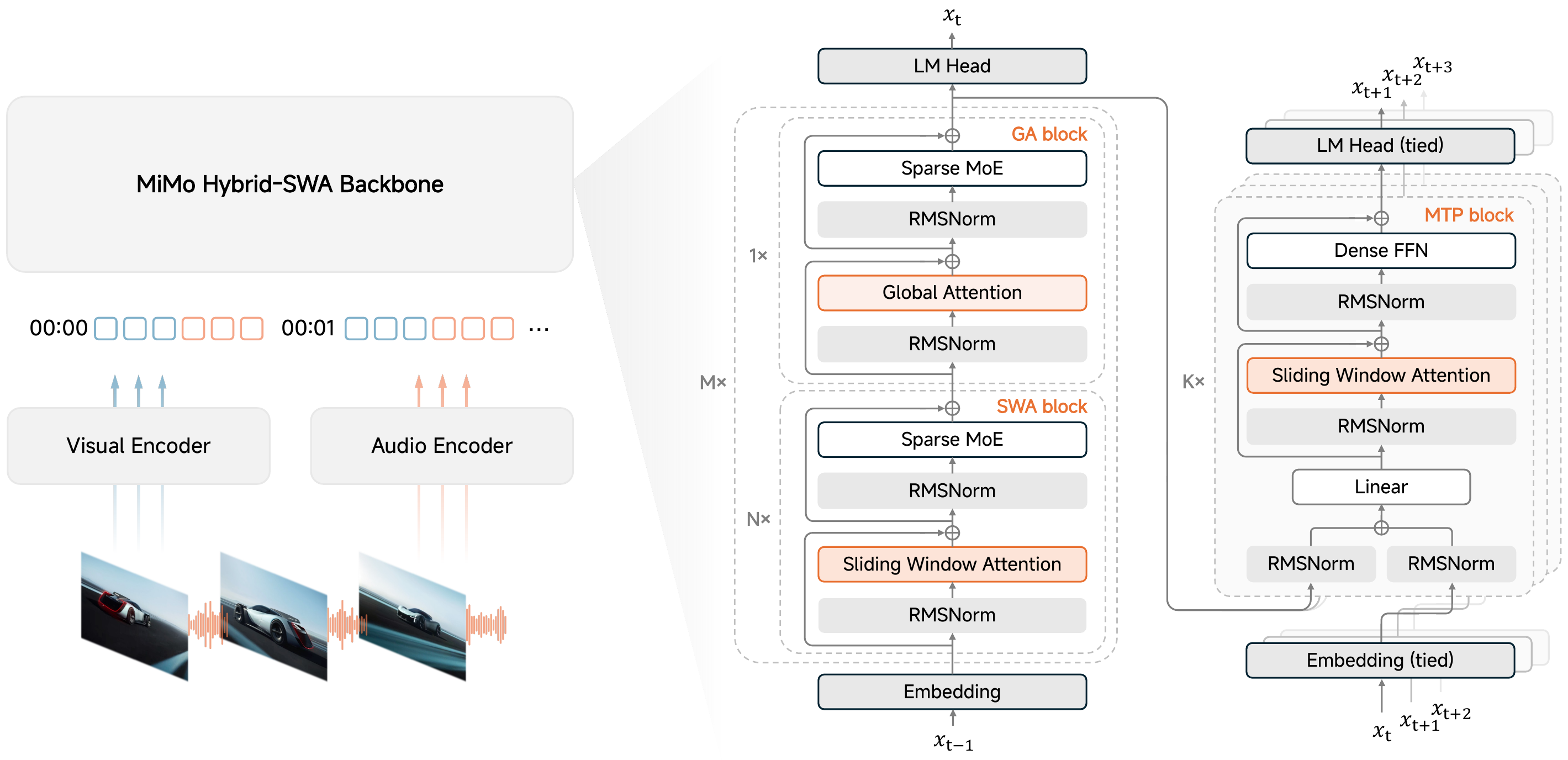}
  \caption{Overall architecture of \mimox{}. Audio, visual, and text inputs are mapped into a shared token sequence and processed by the MiMo Hybrid-SWA backbone, followed by the language-modeling head and multi-token prediction (MTP) blocks.}
  \label{fig:overall-architecture}
\end{figure}

As illustrated in Figure~\ref{fig:overall-architecture}, \mimox{} follows a standard Transformer~\citep{vaswani2017attention} backbone, augmented with visual and audio encoders connected through lightweight projectors.

The text backbone of \mimox{} is mainly composed of repeated hybrid blocks that interleave Local Sliding Window Attention (SWA) and Global Attention (GA).
It stacks $M$ hybrid blocks, each structured with $N$ consecutive SWA blocks followed by a GA block. 
The only exception is the very first Transformer block, which uses global attention with a dense Feed-Forward Network (FFN) to stabilize early representation learning. The sliding window size $W$ used in \mimox{} is $128$. Both the SWA block and the GA block utilize a sparse MoE FFN without shared experts. 
\mimox{} also integrates an SWA and dense FFN based MTP~\citep{gloeckle2024better, liu2024deepseek, xia2025mimo} to improve model performance during pre-training.
A more comprehensive description of the text backbone architecture can be found in \citet{mimo2026v2flash}.
The model configurations are shown in Table~\ref{tab:model_config}.

\begin{table*}[!t]
  \small
  \centering
  \setlength{\tabcolsep}{8pt}
  \resizebox{\textwidth}{!}{
  \begin{tabular}{llcc}
  \toprule
  \textbf{Block} & \textbf{Configuration} & \textbf{MiMo-V2.6-Flash} & \textbf{MiMo-V2.6-Pro} \\
  \midrule

  \multirow{9}{*}{\textbf{Main Block}}
  & Layers (Total/SWA/GA)            & $48 / 39 / 9$ & $70 / 60 / 10$ \\
  & Hidden Size                     & $4096$        & $6144$ \\
  & SWA Heads (Q/KV)                 & $64 / 8$      & $128 / 8$ \\
  & Sliding Window Size             & $128$         & $128$ \\
  & GA Heads (Q/KV)                  & $64 / 4$      & $128 / 8$ \\
  & Head Dimensions (QK/V)           & $192 / 128$   & $192 / 128$ \\
  & Experts (Total/Activated)        & $256 / 8$     & $384 / 8$ \\
  & \# Total Parameters        & $310$B     & $1.02$T \\
  & \# Active Parameters        & $15$B     & $42$B \\
  \midrule

  \multirow{8}{*}{\textbf{MiMo-ViT}}
  & Layers (Total/SWA/GA)            & \multicolumn{2}{c}{$28 / 24 / 4$} \\
  & Hidden Size                     & \multicolumn{2}{c}{$1280$} \\
  & Attention Heads (Q/KV)           & \multicolumn{2}{c}{$32 / 8$} \\
  & Head Dimension                  & \multicolumn{2}{c}{$64$} \\
  & Patch Size ($T \times H \times W$)
                                    & \multicolumn{2}{c}{$2 \times 16 \times 16$} \\
  & Sliding Window Size (Left/Right)
                                    & \multicolumn{2}{c}{$64 / 64$} \\
  & Spatial Merge Size              & \multicolumn{2}{c}{$2 \times 2$} \\
  & \# Parameters                   & \multicolumn{2}{c}{$681$M} \\
  \midrule

  \multirow{8}{*}{\shortstack[l]{\textbf{Audio Tokenizer Encoder}}}
    & Layers (Total/SWA/GA)            & \multicolumn{2}{c}{$24 / 12 / 12$} \\
    & Hidden Size                     & \multicolumn{2}{c}{$1024$} \\
    & Attention Heads (Q/KV)          & \multicolumn{2}{c}{$16 / 16$} \\
    & Head Dimension                  & \multicolumn{2}{c}{$64$} \\
    & Mel Bins                        & \multicolumn{2}{c}{$128$} \\
    & Sliding Window Size
                                      & \multicolumn{2}{c}{$128$} \\
    & Codebooks                       & \multicolumn{2}{c}{$20$} \\
    & \# Parameters                   & \multicolumn{2}{c}{$308$M} \\
    \midrule

  \multirow{6}{*}{\textbf{Audio Patch Encoder}}
  & Layers                          & \multicolumn{2}{c}{$6$} \\
  & Hidden Size                     & \multicolumn{2}{c}{$1024$} \\
  & Attention Heads (Q/KV)           & \multicolumn{2}{c}{$16 / 16$} \\
  & Head Dimension                  & \multicolumn{2}{c}{$64$} \\
  & Attention Group Size            & \multicolumn{2}{c}{$4$} \\
  & \# Parameters                   & \multicolumn{2}{c}{$127$M} \\
\midrule

\multirow{6}{*}{\textbf{Speculative Decoder}}
  & Layers (Total/SWA/GA) & $5 / 5 / 0$ & $5 / 5 / 0$ \\
  & Hidden Size            & $4096$ & $6144$ \\
  & SWA Heads (Q/KV)       & $64 / 8$ & $128 / 8$ \\
  & Sliding Window Size   & $1024$ & $1024$ \\
  & GA Heads (Q/KV)        & $64 / 4$ & $128 / 8$ \\
  & Head Dimensions (QK/V) & $128 / 128$ & $128 / 128$ \\
\bottomrule
  \end{tabular}}
  \caption{Detailed model configurations of MiMo-V2.6-Flash and MiMo-V2.6-Pro.
  Main block layer counts exclude MTP modules.
  Encoder parameter counts include input embeddings but exclude projectors. The audio tokenizer encoder parameter count excludes EMA codebooks.}
  \label{tab:model_config}
  \end{table*}

\subsection{MiMo-ViT}
MiMo-ViT adopts a hybrid attention architecture. It replaces fixed, non-overlapping window attention in MiMo-VL-7B~\citep{yue2025mimo} with sink-augmented SWA, enabling information exchange across window boundaries over successive layers and mitigating visual fragmentation. Local SWA layers alternate between row-major and column-major token serialization to support information propagation along both spatial axes. Finally, GA layers are inserted periodically to aggregate global context directly. This design substantially reduces the computational cost of high-resolution visual processing while achieving performance comparable to GA ViTs. Please refer to \citet{he2026semantic} for more analysis.
To pre-train MiMo-ViT from scratch, we pair it with a small pre-trained LLM and optimize the resulting VLM solely on multimodal understanding data using a cross-entropy objective. This simple recipe avoids auxiliary objectives such as contrastive learning, making pre-training efficient and scalable. Crucially, the LLM remains trainable to provide stable, semantically meaningful gradients to the ViT. After pre-training on more than 4T image tokens, MiMo-ViT learns strong visual representations and serves as a robust foundation for MiMo-V2.6.

\subsection{Audio Encoder}
Audio encoding in \mimox{} consists of two stages: audio tokenization with the Audio Tokenizer, followed by patch encoding. The Audio Tokenizer encoder first processes log-mel spectrograms through a two-layer convolutional frontend that halves the
frame rate. The resulting sequence is then passed through a Transformer with a causal hybrid attention architecture that interleaves SWA and GA layers. SWA layers capture local temporal dependencies, while GA layers aggregate context across the full
preceding sequence. A subsequent downsampling convolution further halves the frame rate to $25$~Hz, after which a 20-layer residual vector quantizer (RVQ) represents each frame as $20$ discrete audio tokens. The Audio Tokenizer follows the training
recipe of MiMo-Audio~\citep{xiaomi2025mimo} and is trained on 20 million hours of audio spanning speech, music, and other audio content.

The audio patch encoder follows the architectural design of MiMo-Audio and is trained jointly with the text backbone. At each time step, the audio tokens are embedded using separate embedding tables, one per RVQ codebook, and the resulting embeddings
are summed to form a single frame representation. Every four consecutive frames are grouped into an audio patch and processed by a Transformer with bidirectional self-attention confined to that patch. The four output representations are then
concatenated and linearly projected into a single backbone input embedding, reducing the audio sequence rate from $25$~Hz to $6.25$~Hz.

\subsection{Speculative Decoder}

Speculative decoding in \mimox{} uses a multi-token prediction (MTP) module following the block diffusion design of DFlash~\citep{chen2026dflash}. The drafter comprises 5 Transformer layers with dense feed-forward networks. Conditioned on backbone hidden features and a clean anchor token, it predicts 7 subsequent tokens in a single forward pass for parallel verification by the backbone.
All draft layers use sliding-window attention (SWA) with grouped queries to limit attention computation and KV-cache size. Tokens attend bidirectionally within their draft block and to at most $1{,}024$ backbone context positions preceding the anchor.

%% file: 3-pretrain.tex
\section{Pre-Training}

\subsection{Pre-Training Setup}

The pre-training corpus of \mimox{} spans text, vision, and audio. The text corpus draws from a broad range of sources, including public web content, books, academic papers, code, and STEM materials.
The vision corpus includes image-captioning, grounding, OCR, GUI, conversation, video, and visual-coding data.
For audio, we curate diverse, high-quality data at scale and organize the data into three task formats: speech-text interleaving, automatic speech recognition (ASR), and general audio captioning.

We adopt a two-stage pre-training strategy. In the first stage, we train the language backbone on text-only data to establish strong foundational language capabilities. In the second stage, we integrate the backbone with our in-house pre-trained ViT and audio encoder and jointly train the full model on omni-modal data, enabling it to understand images, videos, and audio.

We begin pre-training with a context length of 32K tokens and extend it to 256K partway through training. 
\mimox{}-Flash is trained on 48T tokens, comprising 26T tokens in the text stage and 22T in the omni stage. \mimox{}-Pro is trained on 30T tokens, with 27T and 3T tokens in the two stages, respectively.
We employ the AdamW optimizer in the pre-training of \mimox{}.

\subsection{Mid-Training Setup}

Pre-training builds broad knowledge and general understanding across text, vision, and audio. Mid-training bridges this general foundation and subsequent large-scale RL by further developing the model's agentic abilities, extending long-context support, and adapting the optimization setup for large-batch training.

To this end, we train the \mimox{} model series on a diverse, agent-centric data mixture. Its diversity spans both task domains and data modalities, combining realistic agent trajectories across coding, general, visual, and research tasks with high-quality text, repository-level code, and image, video, and audio data.
Training proceeds in two stages: we first train with a 256K context length, allocating the majority of compute to this stage, and then extend the context length to 1M in the final stage.

Our base model was pre-trained with AdamW, but our preliminary experiments showed diminishing optimization efficiency as the batch size increased in mixed-task RL.
Unlike AdamW, which adapts parameters element-wise, Muon leverages the matrix structure of hidden weights by orthogonalizing their updates~\citep{jordan2024muon}. This matrix-based update retains stronger data efficiency beyond the critical-batch-size regime, making Muon particularly attractive for large-batch training~\citep{liu2025muon,shah2025practicalmuon}. 
We therefore switch to a variant of Muon optimizer, Muown, for the hidden weight matrices during mid-training to prepare the model for subsequent large-batch RL. Muown augments Muon with explicit row-norm control, mitigating spectral-norm drift and reducing sensitivity to weight decay at negligible overhead~\citep{lion2026muown}. Embeddings, the LM head, and the MoE router continue to use AdamW.

Prior work reports that switching an Adam-pre-trained model to Muon training can cause optimizer mismatch and degraded performance~\citep{qu2026muonfinetune}.
Nevertheless, during our mid-training, we observe no loss spike throughout the training process.

We employ MXFP4 quantization-aware training (QAT) during mid-training, allowing the model to adapt to low-precision computation while preserving model quality.

%% file: 4-rlscale.tex
\section{Scaling Reinforcement Learning}
In this release, we continue to push the boundaries of post-training algorithms, with a particular focus on scaling RL compute.
Following a short supervised fine-tuning (SFT) stage, we scale RL along three dimensions, training computation~(\S\ref{subsec:rl_compute}), environments and agent harnesses~(\S\ref{subsec:posttrain_data}), and grader computation~(\S\ref{subsec:grader_compute}), achieving a fundamental leap in model capabilities.

\subsection{Scaling RL Training Computation}
\label{subsec:rl_compute}

\begin{figure}[t]
    \centering
    \includegraphics[width=\linewidth]{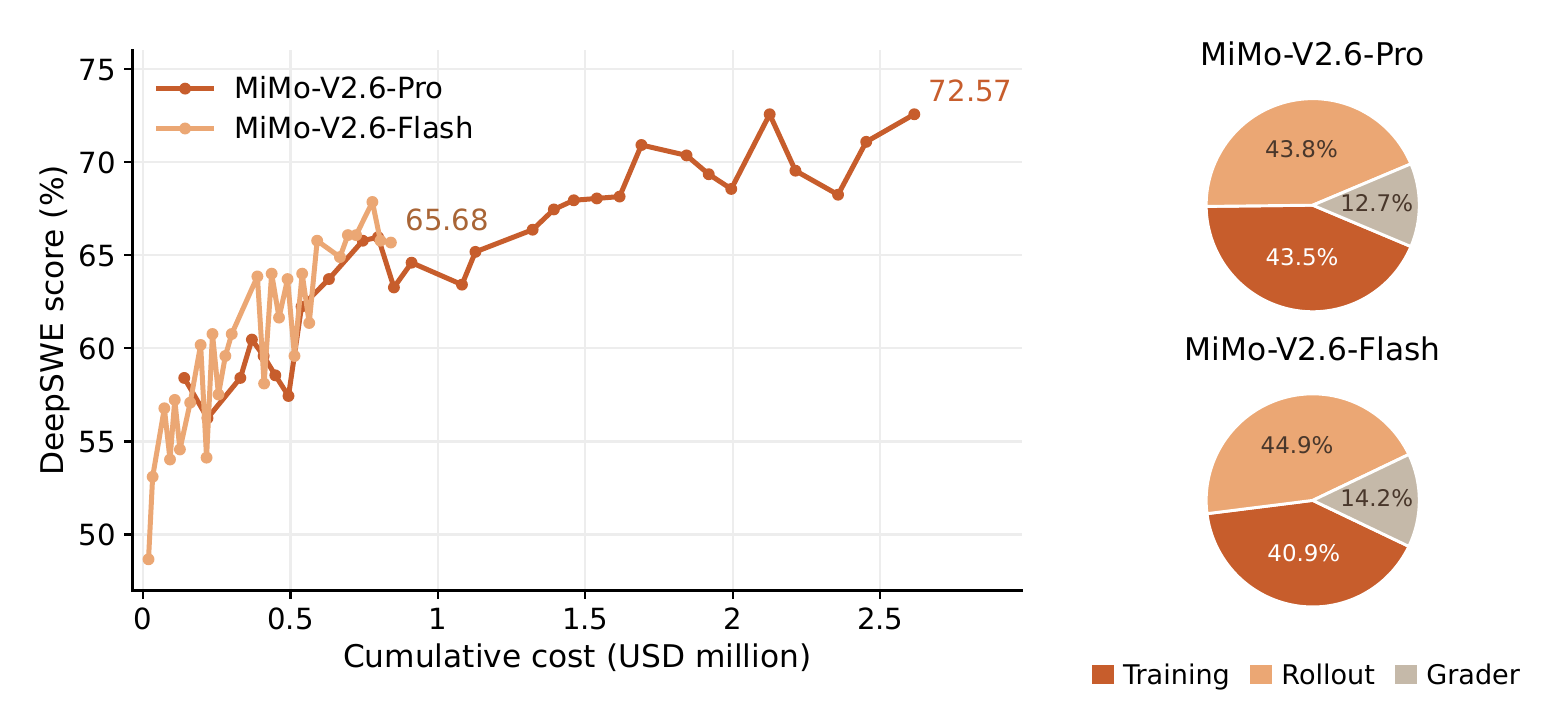}
    \caption{Left: average@3 score on DeepSWE v1.1 versus cumulative RL cost. Right: training, rollout, and grader cost shares for MiMo-V2.6-Pro and MiMo-V2.6-Flash.}
    \label{fig:deepswe_cost}
    \label{fig:deepswe_cost_breakdown}
\end{figure}

We scale RL computation across thousands of GPUs in a single run, spending \$2.6M and \$0.9M on RL post-training for \pro{} and \flash{}, respectively.
We use a large training batch: 1{,}568 prompts with group size $G{=}16$, so each step rolls out 25K sequences that amount to 2.7B--3.7B training tokens (roughly 110K--150K tokens per sequence).
The left panel of Figure~\ref{fig:deepswe_cost} shows the benefit of scaling RL computation.
The average@3 score on the DeepSWE benchmark~\citep{datacurve2026deepswev11} rises steadily with cumulative cost for both models: over the course of RL, \pro{} improves from $58.4$ to $72.6$ and \flash{} from $48.7$ to $65.7$.
The computation splits into three parts: rollout, grading, and training.
Such scaling calls for solid infrastructure to keep training efficient and stable.

We form the RL learning objective as follows:
\begin{equation}
\label{eq:rl_formula}
\mathcal{L}(\theta) = -\mathbb{E}_{q\sim\bigcup_d\mathcal{D}_d,\;\{o_i\}_{i=1}^{G}\sim\mu_{\theta_{\mathrm{old}}}(\cdot\mid q)}\left[\frac{1}{\sum_{i=1}^{G}|o_i|}\sum_{i=1}^{G}\sum_{t=1}^{|o_i|} r_{i,t}\,M_{i,t}\,A_i\,\log\pi_{\theta}(o_{i,t}\mid q,o_{i,<t})\right],
\end{equation}
where $r$ denotes the importance sampling ratio, $M$ is the token-level mask, and $A$ is the advantage. The computation can be decomposed into three parts.
The first is rollout: for every prompt $q$ drawn from the union of task datasets $\bigcup_d \mathcal{D}_d$, the rollout policy $\mu_{\theta_{\mathrm{old}}}$ generates a group of $G$ candidate solutions $\{o_i\}$.
The second is grading: we invest additional compute in agentic evaluation to distinguish effective solutions and behaviors from flawed ones, beyond what binary test outcomes reveal. These judgments inform the advantages $A_i$, enabling more accurate credit assignment across trajectories.
The third is training: the collected tokens update $\theta$ through the gradient of $\log\pi_{\theta}(o_{i,t}\mid q,o_{i,<t})$, with prompt-mean aggregation.
The right panel of Figure~\ref{fig:deepswe_cost_breakdown} breaks down the cost of \pro{} by where the compute goes: rollout consumes 43.8\% and training 43.5\%, while the grader takes the remaining 12.7\%.
A large batch draws from diverse RL tasks spanning various environments and harnesses within one run (Section~\ref{subsec:posttrain_data}).
We also scale the grader to provide fine-grained learning signals (Section~\ref{subsec:grader_compute}).

A large batch is good for scaling out GPUs: rollout parallelizes over sequences and training shards the batch across data-parallel ranks, so throughput grows with the GPU count.
With HBM mostly occupied by the running batch, we can keep a high arithmetic intensity to fully use the GPU in rollout.
Since rollout and grading have a long tail, we use partial rollout~\citep{team2025kimi} to keep the running batch saturated: when we collect a training batch, we leave the in-flight sequences interrupted and resume them in the next rollout phase.
The cost is re-prefill: a continuation must rebuild its KV cache after each policy update, so batch size and partial rollout are chosen together---a large batch amortizes the re-prefill.
To support stable RL training, train--inference consistency is ensured by R3~\citep{ma2025stabilizing} and the replay of top-\textit{p} sampling candidate sets~\citep{liu2025deepseek,mai_thinking_1}.
To ensure that all samples contribute to training with effective gradients, we incorporate a dynamic sampler~\citep{yu2026dapo} to filter groups that are all-pass or all-fail.
In multi-task RL, we implement a \textit{Sample Mixer} to ensure efficient and stable training batch distribution given various rollout durations and pass rates of different tasks.
To pack the trajectories from rollout into large training batches (2.7B--3.7B tokens), we decouple the data plane and control plane.
The infrastructure behind these designs is described in Section~\ref{sec:rlinfra}.

\subsection{Scaling RL Environments and Harnesses}
\label{subsec:posttrain_data}

Scaling agentic RL requires diverse environments that reflect real-world tasks, support reproducible execution at training scale, and provide rewards aligned with task objectives. Meeting these requirements jointly is challenging: realistic workflows involve heterogeneous tools and complex state, while automated graders may be incomplete, inconsistent, or vulnerable to exploitation. We therefore invest in large-scale environment synthesis and curation across coding~(\S\ref{subsubsec:code_task}), general professional workflows~(\S\ref{subsubsec:general_task}), visual artifacts~(\S\ref{subsubsec:design_task}), and cybersecurity~(\S\ref{subsubsec:cyber_task}), emphasizing supervision quality. 
This subsection describes how we construct tasks and environments across these domains, the harnesses through which agents interact with them~(\S\ref{subsubsec:multi_harness}), and the execution checks, rollout-based audits, and adversarial testing used to improve reward reliability and mitigate reward hacking before and during training~(\S\ref{subsubsec:hack_mitigate}).

\subsubsection{Code Agent Tasks}
\label{subsubsec:code_task}

\begin{figure}[t]
    \centering
    \includegraphics[width=\linewidth]{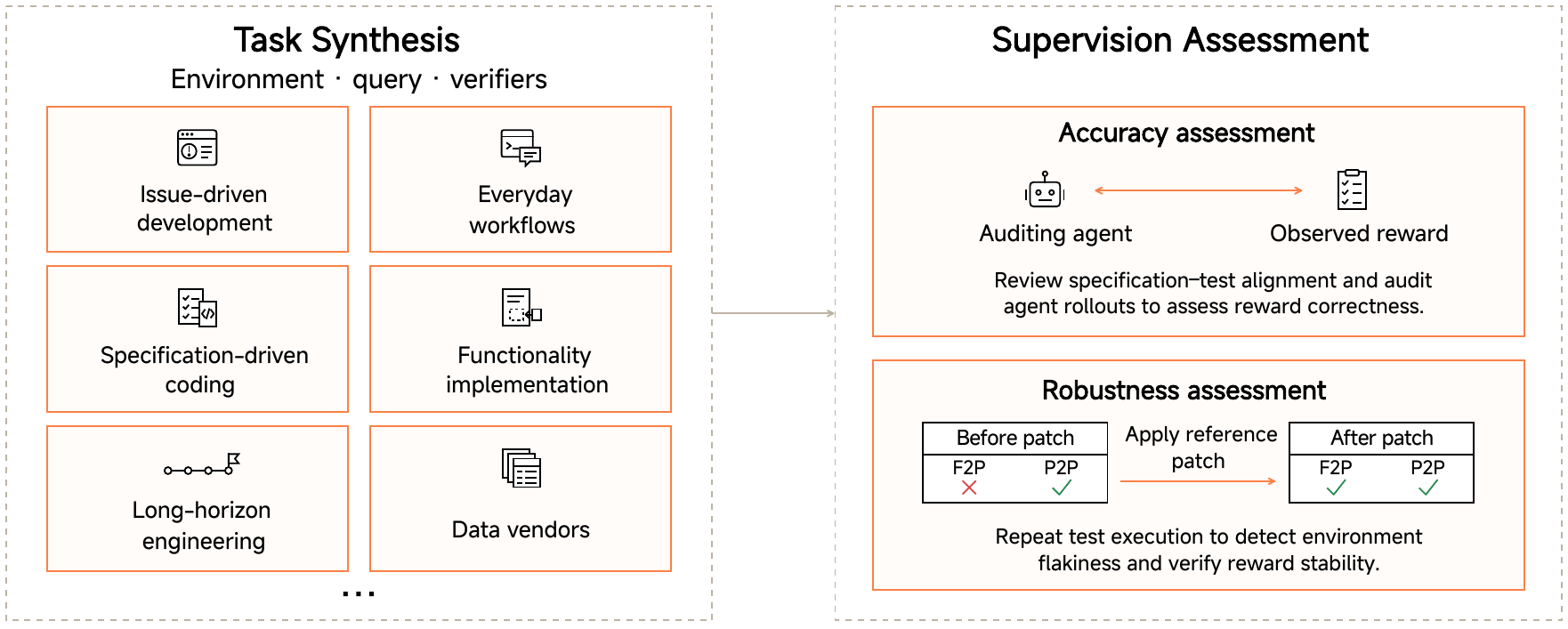}
    \caption{The overall code agent task scaling pipeline. We build tasks from diverse sources and go through rigorous assessment stages to ensure the supervision is accurate and robust.}
    \label{fig:placeholder}
\end{figure}

Coding tasks are particularly well suited to RL, as candidate solutions can be executed and evaluated through automated tests, providing feedback at scale. 
However, executable evaluation alone does not guarantee reliable supervision: tests may overlook required behavior, reject valid implementations, or yield inconsistent outcomes across executions. The central challenge is therefore to scale task construction while ensuring that rewards faithfully reflect task correctness. Motivated by this, we devote substantial effort to building a scalable synthesis and curation pipeline for coding-agent tasks, with an emphasis on the accuracy and robustness of supervision. Figure~\ref{fig:placeholder} summarizes the task synthesis pathways and the checks used to assess supervision accuracy and robustness.

\paragraph{Scalable Task Synthesis from Diverse Sources}
We scale task construction through five complementary synthesis pathways, designed to cover heterogeneous programming languages, development settings, and task horizons. First, in \emph{GitHub-based synthesis}, we collect pull requests and their associated issues and apply heuristic and model-assisted filters to remove duplicate, malformed, or non-reproducible candidates. When a pull request is linked to an issue that provides a sufficiently complete description of the intended change, we use the issue as the task specification. Otherwise, an LLM reconstructs an issue-style specification from the reference patch and the surrounding repository context, while being explicitly instructed to omit implementation-specific details that could reveal the solution. Second, to capture \emph{everyday development workflows}, including interactive ``vibe coding'' scenarios, we collect real-world development requests contributed by employees within our organization. These requests cover common activities such as feature implementation, refactoring, debugging, and repository maintenance, for which agents generate and implement test cases that operationalize the requested behavior. Third, \emph{specification-driven tasks} emphasize code generation under detailed, multi-constraint requirements, allowing us to evaluate whether models can faithfully translate complex specifications into executable implementations. Fourth, in \emph{source-code-driven synthesis}, we use CodeMidas~\citep{ye2026codemidas} to derive tasks from functionality implemented in existing codebases. Agents identify candidate functionality and translate its observable behavior into task specifications and executable environments, this pathway enables scalable task construction across diverse repositories without requiring issues, pull requests, or other development artifacts. For \emph{long-horizon software-engineering tasks}, agents iteratively elaborate task requirements, expand the scope of the required changes, introduce dependencies across components, and construct tests that exercise the resulting multi-step behavior.
We also include and filter samples from public sources~\citep{swelego,badertdinov2026swerebenchv2languageagnosticswe,zan2025multiswebench,zhao2026immersiongithubuniversescaling,yang2026terminallego,swepi} and licensed data vendors to further enrich the task distribution. Collectively, these pathways produce tasks spanning a broad range of programming languages, development contexts, implementation complexities, and task horizons.

\paragraph{Accuracy of Supervision}
We align the behavioral scope of each task specification with that of its unit tests. The goal is to accept implementations that satisfy the specification and reject those that violate it, without requiring incidental details of the reference implementation. We review tasks for insufficient coverage and overly restrictive checks. In particular, tests that enforce requirements absent from the specification are revised or removed.
As inspection alone may miss discrepancies between specifications and tests, we further assess their alignment through rollout-based auditing. Each task is attempted four times by a coding agent. An auditing agent receives all four rollouts in a shared workspace containing the problem statement, tests, reference patch, and each rollout's submitted patch, test output, and full conversation log. The auditor first articulates what a correct solution requires and assesses whether the tests capture those requirements. It then evaluates each submitted solution against the specification using its patch and conversation log, and compares this assessment with the observed reward. A passing solution judged incorrect is flagged as a potential false positive, suggesting incomplete verification. A failing solution judged correct is flagged as a potential false negative, suggesting overly restrictive tests or execution failures. These disagreements identify tasks requiring further review and provide concrete evidence of possible mismatches between the intended behavior and the implemented verification.

\paragraph{Robustness of Supervision}
We check reward stability through repeated execution. For those tasks with reference patch, we ensure fail-to-pass (F2P) tests must fail and pass-to-pass (P2P) tests must pass before applying the reference patch; after the patch, both must pass. We require these outcomes to remain stable across eight reruns, screening for flaky tests and environment-induced reward fluctuations.
We also reuse the previous rollout pipeline to collect agent trajectories to investigate potential reward hacking. The submitted patches and conversation logs provide evidence of how agents obtained their rewards, helping identify unintended shortcuts or exploitation of the environment and evaluation process. These checks complement the assessment of solution correctness and inform the broader reward-hacking mitigation procedures described below.

Together, these procedures support the construction of a diverse coding-task corpus under a consistent standard of supervision quality. Specification--test alignment targets the fidelity of the reward signal, repeated execution checks its stability, and rollout-based auditing probes whether it agrees with an independent assessment of solution correctness. Our objective is to make this combination scalable, so that RL training can benefit from broad task coverage while encouraging agents to satisfy the intended requirements rather than exploit gaps in evaluation.
\subsubsection{General Agent Tasks}
\label{subsubsec:general_task}

 \begin{figure}[t]
      \centering
      \includegraphics[width=0.9\linewidth]{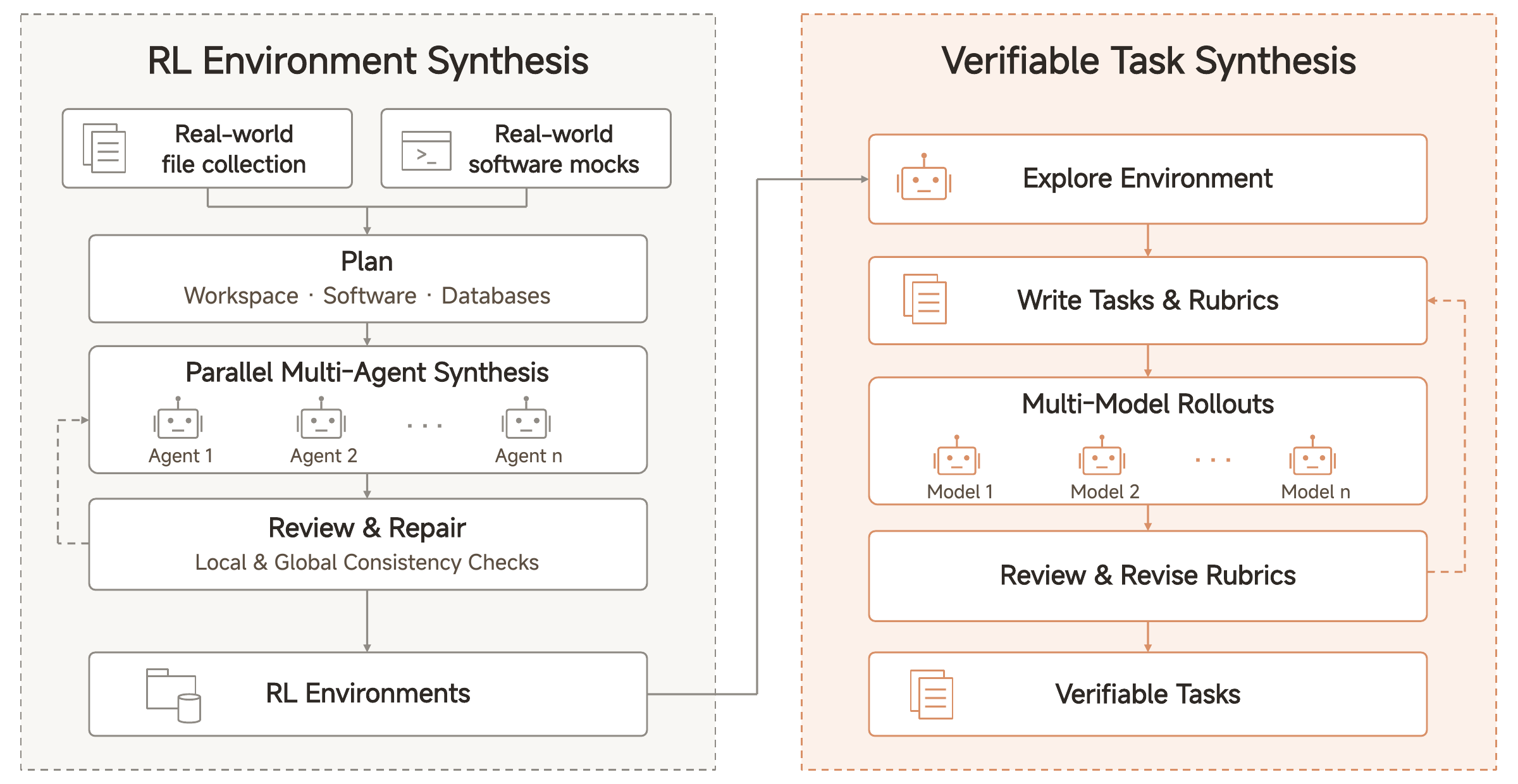}
      \caption{The overall general-agent environment and verifiable-task synthesis pipeline. We first construct realistic, resettable environments from real-world files and software mocks. Agents then
      explore these environments and synthesize tasks with verifiable rubrics.}
      \label{fig:general-agent-synthesis}
  \end{figure}

We extend agent capabilities to real-world professional workflows across diverse domains, whose workflows require agents to analyze heterogeneous documents, use specialized software, and produce deliverables that meet domain-specific standards. We synthesize environments that support such workflows, then construct challenging tasks with verifiable outcomes. Figure~\ref{fig:general-agent-synthesis} summarizes this pipeline, including environment construction, task synthesis, and iterative rubric refinement.

\paragraph{Environment Design} A general-agent environment typically consists of a workspace and a set of software tools, accessible through interfaces such as MCP, APIs, CLIs, and GUIs. Our design balances realism with the requirements of large-scale RL: files, software behavior, and underlying data should reflect professional practice, while execution must remain local and environments must be easy to reset. We therefore collect real-world files for direct inclusion or as references for synthesis, and use an automated multi-agent workflow to build local software mocks that reproduce supported operations, state changes, output formats, and error responses. Tools that already operate without external services are integrated directly. All state is stored locally, and each rollout runs in an isolated sandbox that can be restored to a fixed initial state. This avoids network variability and service limits while supporting reproducible execution.

\paragraph{Environment Synthesis} To assemble these components into coherent scenarios, a planning agent specifies the workspace structure, selects appropriate tools, and plans file and database contents. Web search grounds the plan in real-world information, while relationships across files and between files and database records support subsequent multi-hop reasoning and cross-validation. Multiple agents then generate files and populate databases in parallel, following the shared plan and each mock's data specification. A review agent checks individual artifacts and global consistency, including entity names, numerical reconciliation, timelines, and references. Iterative repair resolves inconsistencies before task construction. The workspace and tool configuration vary by scenario, so each environment reflects the relevant working practices without requiring every available file type or software interface.

\paragraph{Task Synthesis and Verification} We survey common professional tasks and combine and generalize them into seed tasks. An agent explores each environment and uses these seeds to construct tasks suited to its contents and tools. Task completion is evaluated using atomic, binary rubric items: code-based checks verify deterministic properties, such as database values and deliverable formats, while LLM-based checks assess more open-ended content. For LLM-based items, agreement across repeated judgments by the same model and across different judge models helps identify ambiguity.

Consistency alone does not establish that rubrics capture task completion. We therefore collect rollouts from models with different capability levels and have a review agent examine their trajectories, execution results, and rubric judgments. The reviewer revises overly restrictive criteria that reject valid solutions and overly permissive criteria that accept incomplete ones. To test resistance to reward hacking, we include negative checks for unintended changes to unrelated files or databases and construct adversarial solutions that appear correct without actually solving the task. During RL, a self-hosted MiMo-V2.6-SFT model serves as the grader to support stable scoring. This process yields thousands of environments and diverse, verifiable tasks, which we combine with other tasks for RL training.

\subsubsection{Visual Agent Tasks}
\label{subsubsec:design_task}

To empower agents to shape the digital world through creative expression and precise visual control, we curate a diverse suite of visual tasks in which agents work with a broad range of artifacts, such as websites, interactive applications, games, 3D scenes, slides, SVG, videos and Figma designs. We organize these tasks into two complementary categories: open-ended design and high-fidelity visual replication. Together, they develop the model's ability to produce reliable implementations with strong aesthetic quality and precise visual control.

Open-ended design seeks to produce aesthetically compelling visual artifacts that satisfy user intent. The diversity of valid creative solutions makes aesthetic quality difficult to capture through fixed rules alone. We therefore combine pointwise and groupwise grading to establish consistent quality standards while rewarding relative improvements in aesthetic quality. Specifically, we first iteratively refine pointwise rubrics for runtime correctness, instruction adherence, layout integrity, and basic aesthetic quality. Once these rubrics stabilize, we introduce groupwise grading that jointly compares the rendered artifacts within each rollout group for the same query, identifying clearly stronger and weaker candidates and assigning rewards accordingly.

High-fidelity visual replication aims to faithfully reproduce a specified visual target. Compared to open-ended design, the explicit reference enables more direct evaluation of visual fidelity. We grade these tasks primarily through rule-based similarity metrics, such as pixel-level similarity between rendered artifacts and their references, supplemented by LLM-based judging for holistic visual assessment. Together, these signals support effective scaling of reinforcement learning on visual replication tasks.

\subsubsection{Cybersecurity Agent Tasks}
\label{subsubsec:cyber_task}

We train cyber agents with RL on real-world vulnerability reproduction: given a project and a target bug,
the agent must construct an input that triggers it. This foundational offensive-security skill underlies
exploit development, privilege escalation, and CTF challenges. It is also well suited to RL at scale:
OSS-Fuzz supplies tens of thousands of manually confirmed instances, a volume unmatched by other
security tasks. The
difficulty is not triggering \emph{a} crash, since complex C/C++ projects expose dozens of reachable crash
paths, but triggering the \emph{specific} vulnerability described. Verification must therefore separate the
intended bug from unrelated crashes, and because it doubles as the RL reward, it must be accurate,
deterministic, and cheap.

Existing oracles fall short. \emph{Fix-binary differential testing}, as used in CyberGym~\citep{DBLP:journals/corr/abs-2506-02548}, accepts a PoC that crashes
the vulnerable binary but not the patched one; an incomplete patch rejects correct PoCs, and unrelated
changes between the two commits flip verdicts for reasons unconnected to the bug. Either error corrupts the
training gradient. \emph{LLM-based judging} returns different verdicts for the same PoC across runs and
cannot serve as a stable reward. We instead extract two attributes from the ground-truth sanitizer
(ASan/MSan/UBSan) report: the \emph{vulnerability type} (e.g., heap-buffer-overflow) and the \emph{crash
location} (the topmost project-level stack frame). A PoC is accepted iff its crash matches both under
rule-based string matching, which is deterministic, reproducible, and computationally trivial. The task
description is derived from the same report, stating the exact type and function in which the crash must
occur, so description and verification share a single source of truth; CyberGym's LLM-generated
descriptions, by contrast, are often too broad (e.g., ``buffer overflow in libxml2'') or simply wrong.

For each vulnerability we check out the source at the reported commit, build the fuzzing harness, and give
the agent the complete runtime environment: full source plus the compiled harness binary. CyberGym
withholds the binary, restricting agents to source-only analysis; providing it mirrors real vulnerability
analysis, where a researcher runs the target under a debugger, inspects memory, and crafts mutations from
runtime observations.

\subsubsection{Multi-Harness Training}
\label{subsubsec:multi_harness}
Agent harnesses differ in module design and are selected or built to meet different application needs. Models must therefore adapt to diverse interaction mechanisms, including those of harnesses unseen during training, making cross-harness generalization an important capability.

Training exclusively within a single harness may couple task-solving strategies to harness-specific implementations and limiting transfer. We therefore treat harness diversity as an additional training dimension alongside diversity in tasks and environments. A natural approach is to train on several production harnesses such as MiMo Code and Codex. However, this is a poor fit for RL in practice. First, production harnesses wrap the agent loop with engineering safeguards and multi-step workflows, and steer the model with numerous constraint and instruction prompts. These extras fall outside the task-completion reward signal, making credit assignment unreliable and leaving reward-unmeasured requirements to simply be ignored. Second, their modules are tightly coupled rather than independently configurable, so one cannot vary a single interaction mechanism in isolation or assemble diversity by controlled recombination.

To this end, we propose multi-harness training with carefully designed mini-harnesses. All mini-harnesses start from the same minimal agent loop, which already includes the components needed for task completion—system prompt, tools, and context management. These modules stay minimal and decoupled, so implementations can be freely recombined and harness expansion remains controllable, safe, and clean. We then derive diverse, task-adapted mini-harness configurations for Code, General, Visual, and Cyber. The resulting training distribution is diverse yet controllable, supporting analysis of individual harness mechanisms and encouraging transferable task-solving strategies. 

\subsubsection{Reward Hacking Mitigation}
\label{subsubsec:hack_mitigate}
\input{tables/hack_patterns}

RL relies on rewards to measure task success, but agents can sometimes earn high rewards by exploiting the environment or the evaluation process. This behavior, known as reward hacking, can be reinforced during training, increasing scores without improving task performance.
For common coding agent tasks like repository-repair, a recurring failure mode is \emph{solution
leakage}: agents obtain a published fix beyond the intended task context and
use it to construct a patch. Table~\ref{tab:reward_hacking_patterns} links each
task request to the agent's stated intent and subsequent action across five
repositories. Such behavior can satisfy the test-based reward without
demonstrating that the agent derived the repair from the bug report and the
assigned checkout.
Followingly, we introduce our mitigation combining mid-training alignment data with RL environment preparation, adversarial screening, and auditing throughout training.

\paragraph{Mid-Training Alignment Data}
In early experiments, we observed a tendency toward reward hacking in MiMo. To mitigate this behavior, we synthesized a set of training examples from these cases and included them in mid-training. For each case, MiMo reflects on the faulty reasoning, revises the relevant turn, and continues with actions grounded in the task specification. The revised reasoning keeps the original error recognizable and makes the correction explicit. We found that adding these examples improved the model's alignment.

\paragraph{Environment Preparation} In coding tasks, agents may pass tests by recovering leaked solutions from the environment or retrieving existing solutions over the network. Building a project with the reference patch applied during environment construction can leave behind artifacts that reveal the solution. We therefore remove build logs, verifier outputs, residual patches, and project-generated binaries or bytecode that could expose it. We also clean caches that may contain solution information, including those outside the repository, while preserving third-party dependencies needed for offline rebuilding. For every task, we retain Git history up to and including the base commit, removing later commits and their associated references. We enforce container-level network isolation to restrict access to upstream fixes and alternative package versions that may contain the solution. These safeguards are accompanied by explicit instructions against retrieving existing solutions or bypassing the required implementation.

\paragraph{Hack Agent} A dedicated hack agent then probes the prepared environments for remaining leaks and exploitable weaknesses. We guide its search with examples from early experiments, including recovering solutions from cached artifacts or preinstalled copies of the target project. The agent checks these known routes while searching for new ones. It uncovered many exploit paths that we had not observed during training and that our existing cleanup procedures did not cover. We use these findings to refine cleanup and access restrictions, then rerun the hack agent on the updated environments. These subsequent checks repeatedly exposed further weaknesses, requiring additional rounds of cleanup and testing. We continued this process until the hack agent could no longer find a successful exploit in any of the environments.

\begin{figure}[t]
    \centering
    \includegraphics[width=\linewidth]{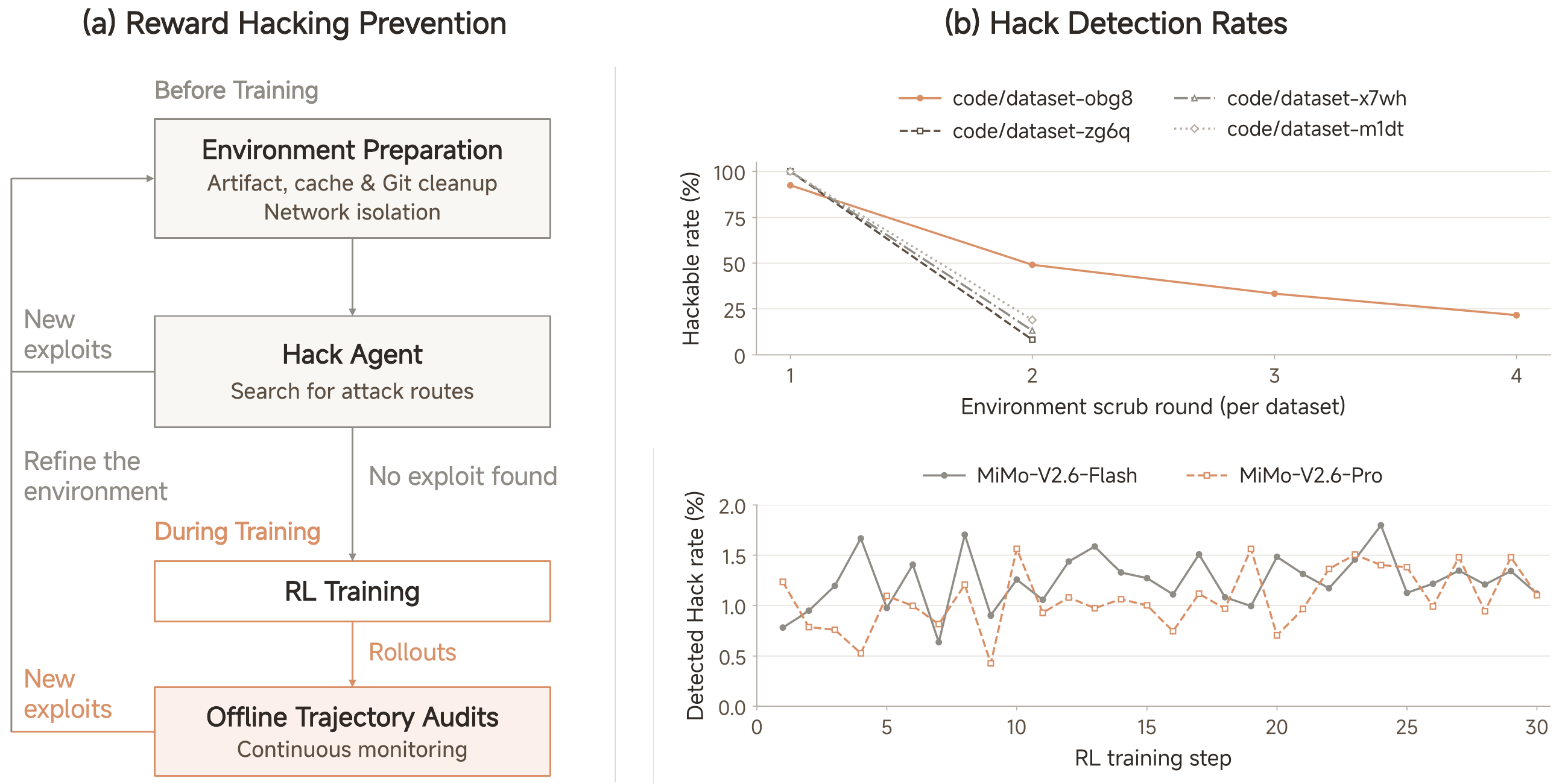}
    \caption{
        Reward hacking prevention and monitoring.
        (a) Environment preparation and iterative hack-agent screening before training, while offline
        trajectory audits monitor hacking behaviors throughout training.
        Findings from both stages are used for environment improvements.
        (b) Top: the fraction of environments found hackable over cleanup rounds.
        Bottom: the fraction of trajectories with detected reward hacking throughout our final RL run.
    }
    \label{fig:reward_hack}
\end{figure}

\paragraph{Training-Time Auditing} We regularly audit agent trajectories offline for reward hacking throughout training. As the policy evolves, it may discover shortcuts that were not found during adversarial screening. We use these observations to identify weaknesses in the environments and guide further cleanup and access restrictions. Alongside these offline audits, the groupwise agentic grader (Section~\ref{subsubsec:groupwise_grading}) sets the effective reward of confirmed hacking trajectories to zero before recomputing group statistics and advantages. With this correction in place, the logged confirmed-hack share remains below $2\%$ throughout the whole training process for both \flash{} and \pro{} (Figure~\ref{fig:reward_hack}).

\subsection{Groupwise Agentic Grading}
\label{subsec:grader_compute}

\begin{figure}[t]
    \centering
    \includegraphics[width=\linewidth]{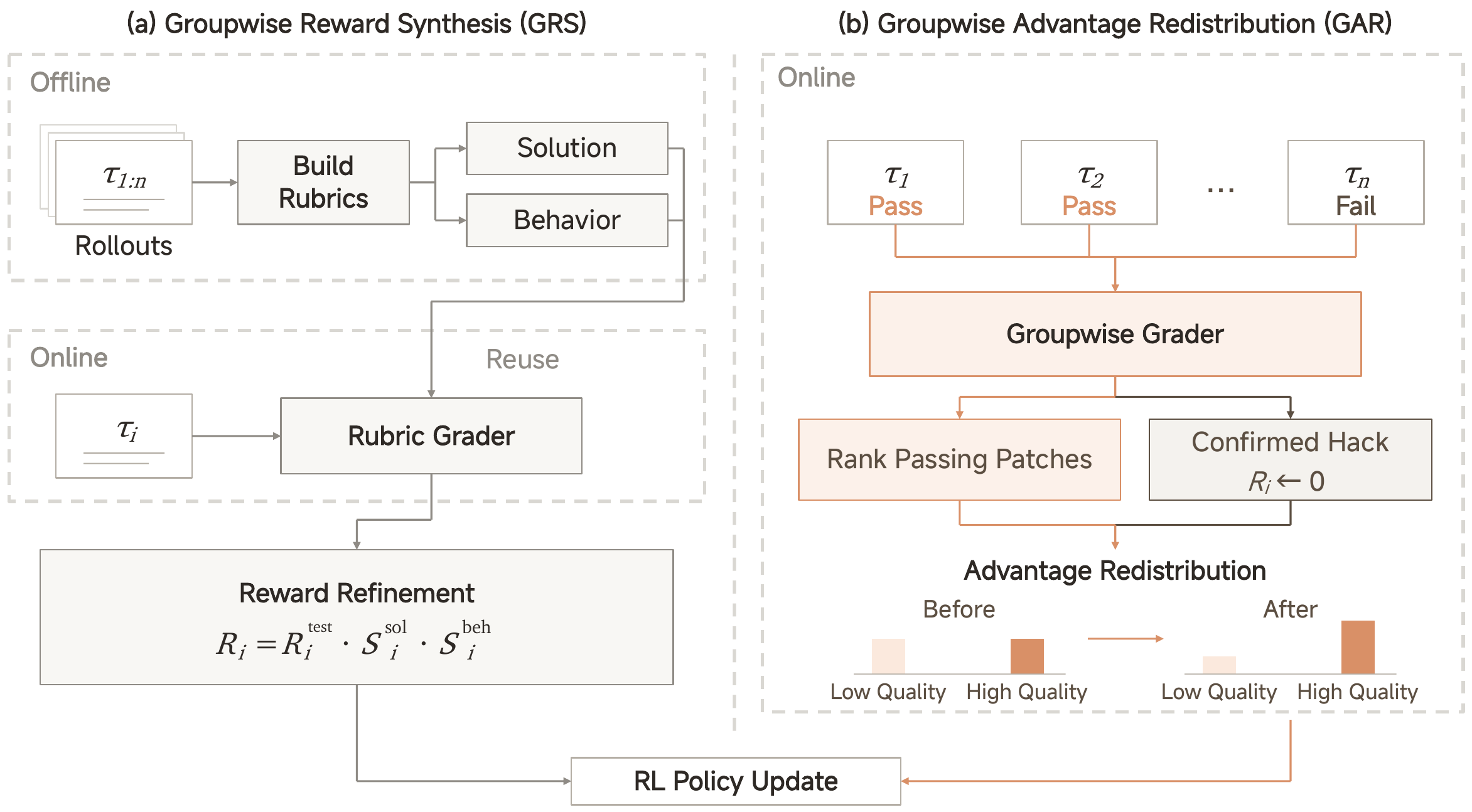}
    \caption{Groupwise agentic grading for code-agent RL. (a) Groupwise reward synthesis combines test outcomes with per-rollout scores from precomputed task-specific rubrics. (b) Groupwise advantage redistribution compares trajectories online, resets confirmed-hack rewards to zero, and redistributes sequence-level advantages. Bars schematically illustrate the redistribution of positive advantage from lower- to higher-quality passing solutions.}
    \label{fig:grader_architecture}
\end{figure}

For code agent tasks, binary test rewards provide a scalable correctness signal but do not distinguish implementation quality or problem-solving behavior among passing solutions. We use two complementary methods on distinct subsets of these tasks. \textbf{Groupwise Reward Synthesis (GRS)} is used for a subset of high-passrate tasks. It compares multiple offline rollouts to construct task-specific rubrics, which are reused during training to score individual rollouts and combine their quality scores with test rewards. For all the remaining code agent tasks, we rely primarily on \textbf{Groupwise Advantage Redistribution (GAR)}~\citep{dong2026groupweise}. Its online agentic grader jointly examines successful and failed trajectories within each mixed-outcome group, ranks passing solutions, and redistributes positive advantage toward higher-quality passing trajectories. Figure~\ref{fig:grader_architecture} summarizes both workflows.

\subsubsection{Groupwise Reward Synthesis (GRS, Offline Rubrics)}

For each selected task, we collect multiple offline rollouts and ask an agent to study them together with the task specification and repository. Comparing these attempts exposes different solution approaches, recurring mistakes, and useful behaviors that may be distributed across several trajectories. The agent turns this analysis into two sets of criteria: solution rubrics, which assess the resulting implementation, and behavior rubrics, which assess how the agent approaches and verifies its work.

Solution rubrics describe task-relevant properties of a good implementation, including satisfaction of the requirements, appropriate handling of edge cases, and changes consistent with the surrounding codebase. Behavior rubrics describe observable practices that support reliable problem solving, such as gathering relevant evidence and checking the effects of code changes. We seek criteria that distinguish meaningful differences in quality while allowing different valid approaches to the task.

The sampled rollouts inform rubric construction, but the criteria are grounded in the task itself. A useful behavior may appear in only one attempt, and a requirement supported by the task may be absent from every observed solution. The agent can therefore identify improvements beyond those demonstrated in the sampled rollouts. At the same time, a choice made by one successful solution does not automatically become a requirement for all others.

The resulting rubrics are reused to evaluate subsequent training rollouts individually. A grader agent enters each rollout's execution environment and assesses it against the task-specific rubrics, using the resulting code, execution results, and trajectory as evidence. It assigns a solution score $S^{\mathrm{sol}}$ for the quality of the implementation and a behavior score $S^{\mathrm{beh}}$ for how the agent approached and verified its work.

For trajectory $i$, let $R_i^{\mathrm{test}}$ denote the original binary test reward. We synthesize the final training reward by multiplying this test reward by the two rubric scores:
\begin{equation}
    R_i =
    R_i^{\mathrm{test}}
    \cdot S_i^{\mathrm{sol}}
    \cdot S_i^{\mathrm{beh}}.
    \label{eq:offline_rubric_reward}
\end{equation}
This multiplicative form keeps rubric supervision tied to test outcomes. Failed trajectories retain zero reward, while passing trajectories are further distinguished by implementation quality and problem-solving behavior. Even when every rollout in a group passes the tests, differences in the product of the two rubric scores can still provide a learning signal. The offline task analysis thus becomes a reusable source of supervision, allowing training to capture quality differences that binary test rewards leave unexpressed.

\subsubsection{Groupwise Advantage Redistribution (GAR, Online Grading)}
\label{subsubsec:groupwise_grading}
We apply online groupwise advantage redistribution to the remaining code agent tasks. For each mixed-outcome rollout group, we place all trajectories in a shared workspace containing the task specification, repository, submitted patches, and test outputs. An SFT-trained agentic grader jointly examines all trajectories within the group, contrasting successful and failed attempts and comparing passing patches along five dimensions: the suitability of the solution approach, precision in implementing that approach without omissions or unnecessary fallbacks, minimality relative to the necessary changes, avoidance of unintended effects outside the task, and craftsmanship consistent with codebase conventions. These assessments guide the groupwise ranking, with ties when differences are inconclusive. The grader can inspect repository code and run targeted tests to better understand the task and verify candidate solutions. When evidence confirms dependence on an external or leaked answer, we reset the trajectory's effective reward to zero and treat it as a failure before recomputing group statistics.

We use the remaining quality rankings to redistribute sequence-level advantages. For trajectory $i$, let $R_i$ be the effective binary reward after hack correction, $\bar R$ its group mean, $A_i=R_i-\bar R$ its sequence-level advantage, and $\mathcal{P}=\{i:R_i=1\}$. For nonempty $\mathcal{P}$, quality factors $f_i\in(0,1]$ first downweight lower-quality passes, after which a common factor redistributes the removed positive advantage mass among passing trajectories. The uncapped update is
\begin{equation}
    \lambda = \frac{\sum_{j\in\mathcal{P}} A_j}
                   {\sum_{j\in\mathcal{P}} f_j A_j},
    \qquad
    A_i' =
    \begin{cases}
        \lambda f_i A_i, & i\in\mathcal{P},\\
        A_i, & i\notin\mathcal{P}.
    \end{cases}
    \label{eq:grader_redistribution}
\end{equation}
This uncapped update preserves $\sum_{i\in\mathcal{P}}A_i'=\sum_{i\in\mathcal{P}}A_i$ and the quality-induced relative weights without altering failed trajectories. In effect, it redistributes positive advantage mass from lower-quality to higher-quality successful trajectories. Simply downweighting positive advantages leaves negative advantages unchanged; renormalization restores their balance as a safeguard against excessive entropy growth. In practice, we cap the common rescaling factor to prevent excessive amplification of positive advantages. We then subtract the group mean from the advantages of both passing and failed trajectories, yielding final sequence advantages $A_i^{\mathrm{new}}$ with zero group mean. We broadcast this sequence-level advantage to all response tokens in the trajectory. Grading runs asynchronously, with unusable grader outputs falling back to the original advantages.

\begin{figure}[t]
    \centering
    \includegraphics[width=\linewidth]{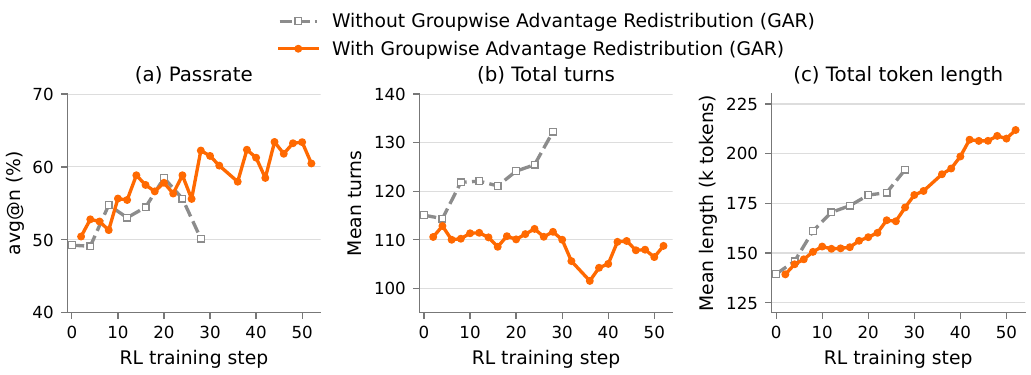}
    \caption{Comparison with and without online groupwise advantage redistribution on DeepSWE v1.1, using code-only RL with \flash{} (training batch size 128). Pass rate is \texttt{avg@n} with $n=3$; we also report the mean total number of turns and total token length.}
    \label{fig:grader_dynamics}
\end{figure}

To assess the effect of online groupwise advantage redistribution on training dynamics, we compare code-only RL runs of \flash{} with and without it, using a training batch size of 128 and token-mean loss aggregation (Figure~\ref{fig:grader_dynamics}). Without online grading, turns and total token length grow rapidly, causing more trajectories to hit the length limit, making it difficult to sustain improvements in pass rate. With online grading, pass-rate gains are sustained through step~52, while turn counts remain roughly stable and token length grows gradually. These trends suggest that online grading supports continued policy improvement under a more stable training regime.

Separate maintainer-oriented audits found that, under pressure to improve test pass rates, policies trained without online grading increasingly adopted undesirable behaviors such as speculative compatibility branches, broad exports, exception swallowing, relaxed validation, and evaluation-specific configuration changes. These workarounds aim to increase the likelihood of passing test but can exceed the scope of the task instructions, expand APIs unnecessarily, obscure failures, and make the code harder to maintain. In contrast, policies trained with online grading tended to produce smaller, more precise patches that remained within the requested scope and were easier to maintain.

\subsubsection{Behavioral Regularization}
\label{subsec:penalty}

To improve RL training stability and encourage efficient, reliable behavior, we introduce two complementary mechanisms. A group-relative length penalty curbs excessive token growth, while segment-level behavioral penalties address format violations and tool call errors. Batch-level advantage rebalancing improves credit assignment while limiting excess negative optimization pressure that can drive uncontrolled entropy growth.

\paragraph{Group-Relative Length Penalty} We find that a group-relative length penalty improves generalization and curbs rapid growth in generated tokens during RL training.
For each prompt $q$ with $G$ sampled rollouts, let $R_i$ be the original outcome reward, $\mathcal P_q$ the indices of successful rollouts, and $\ell_i=|o_i|$ the generated-token count. Let $A\in[0,1]$ be the minimum group pass rate and $B\in(0,100)$ the percentile parameter. For groups with $|\mathcal P_q|/G>A$, we compute the reference length $\ell_q^\star=\operatorname{Quantile}_{B/100}\{\ell_j:j\in\mathcal P_q\}$ and obtain the adjusted reward
\begin{equation}\label{eq:length_penalty}
\widetilde R_i=R_i-\mathbf 1[i\in\mathcal P_q]\,X\left[\operatorname{clip}\!\left(\frac{\ell_i/\ell_q^\star-1-\delta}{s-\delta},\,0,\,1\right)\right]^\gamma .
\end{equation}
Here $X\ge0$ is the maximum reward deduction, $\delta\ge0$ the tolerated relative excess above $\ell_q^\star$, $s>\delta$ the excess at which the penalty saturates, and $\gamma\ge1$ the ramp exponent; $\delta=0$ penalizes successful rollouts longer than the reference length. The indicator $\mathbf 1[\cdot]$ restricts the deduction to successful rollouts, and $\operatorname{clip}(x,0,1)=\min(1,\max(0,x))$.
Other groups retain their original rewards. The outcome advantage $A_i$ in Eq.~\eqref{eq:rl_formula} is computed from the adjusted rewards $\widetilde R_i$; the threshold $A$ is a separate hyperparameter.
This encourages concise successful solutions using a reference adapted to each prompt, while the pass-rate gate preserves room for exploration on difficult prompts.

\paragraph{Segment-Level Behavioral Penalties} Outcome rewards can reinforce faulty intermediate behavior, motivating segment-level rules for format violations and tool call errors such as malformed markup, invalid tool names, or malformed arguments. Let $h_{i,t}=1$ mark flagged tokens and $0$ otherwise. Across the full training batch, $H_\pm$ and $C_\pm$ collect flagged and unflagged token indices $(i,t)$ with loss mask $M_{i,t}=1$, respectively; $\pm$ denotes the sign of the owning trajectory's outcome advantage $A_i$, and all sums below count tokens.
\begin{equation}\label{eq:segment_penalty}
\begin{aligned}\widetilde A_{i,t}&=\begin{cases}\alpha(1-h_{i,t})A_i,&A_i>0,\\\bigl[\beta(1-h_{i,t})+\kappa h_{i,t}\bigr]A_i,&A_i<0,\\0,&A_i=0,\end{cases}\qquad \kappa>1,\\
\alpha&=\min\!\left(\alpha_{\max},1+\frac{\sum_{H_+}A_i}{\sum_{C_+}A_i}\right),\qquad \beta=\max\!\left(\beta_{\min},1-\frac{(\kappa-1)\sum_{H_-}|A_i|}{\sum_{C_-}|A_i|}\right).\end{aligned}
\end{equation}
Here $\kappa>1$ multiplies the magnitude of negative advantage on flagged tokens; $\alpha$ scales unflagged positive tokens upward, while $\beta$ scales unflagged negative tokens toward zero. The hyperparameters $\alpha_{\max}\ge1$ and $0<\beta_{\min}\le1$ cap positive amplification and bound negative attenuation, respectively. The adjusted token advantage $\widetilde A_{i,t}$ replaces $A_i$ in Eq.\eqref{eq:rl_formula}.
This masks flagged tokens in positive trajectories and penalizes them more strongly in negative trajectories. Removed positive advantage is redistributed to unflagged positive tokens, while added negative magnitude is offset by reducing penalties on unflagged negative tokens: behavior without detected errors receives more reinforcement or less punishment, depending on the trajectory's sign.
Each sign's total advantage mass is conserved when neither scale is clipped, limiting excess negative pressure that can drive uncontrolled entropy growth. If a denominator is zero, its scale is set to one; conservation is not guaranteed in this case or when clipping occurs.

%% file: tables/hack_patterns.tex
\begin{table*}[t]
  \centering
  \small
  \definecolor{HackTask}{HTML}{245580}
  \definecolor{HackThinking}{HTML}{6B3C78}
  \definecolor{HackAction}{HTML}{226347}
  \setlength{\tabcolsep}{5pt}
  \renewcommand{\arraystretch}{1.15}

  \begin{tabularx}{\linewidth}{@{}
    >{\raggedright\arraybackslash}p{0.30\linewidth}
    >{\raggedright\arraybackslash}X@{}}
    \toprule
    \textbf{Pattern and shortcut} & \textbf{Illustrative case} \\
    \midrule

    \textbf{Install and read}
    \par Uses a newer release of the target package as an answer key,
    copying a fix absent from the assigned checkout.
    & \textcolor{HackTask}{\textbf{Task (pytest):} Fix Windows \texttt{conftest.py}
      imports broken by path lowercasing.}
      \par\textcolor{HackThinking}{\textbf{Thinking:} ``Let me check the pytest
      changelog or GitHub to see if there's a more recent fix.''}
      \par\textcolor{HackAction}{\textbf{Action:}
      \texttt{pip install pytest==5.4.3}; inspect the installed source.} \\
    \addlinespace[0.65em]

    \textbf{Fetch upstream source}
    \par Downloads an upstream file or patch that already contains the fix,
    exposing the code changes needed to pass the tests.
    & \textcolor{HackTask}{\textbf{Task (Astropy):} Correct the misleading error
      when a required \texttt{TimeSeries} column is removed.}
      \par\textcolor{HackThinking}{\textbf{Thinking:} ``Let me just look at the
      file directly from the GitHub raw URL.''}
      \par\textcolor{HackAction}{\textbf{Action:}
      \texttt{curl .../astropy/timeseries/core.py}} \\
    \addlinespace[0.65em]

    \textbf{Clone upstream}
    \par Reads a newer upstream checkout to reconstruct the published fix
    missing from the assigned historical commit.
    & \textcolor{HackTask}{\textbf{Task (Matplotlib):} Stop \texttt{ax.clear()}
      from restoring hidden ticks and labels on shared axes.}
      \par\textcolor{HackThinking}{\textbf{Thinking:} ``Let me directly fetch
      and inspect the relevant files from the latest matplotlib.''}
      \par\textcolor{HackAction}{\textbf{Action:}
      \texttt{git clone .../matplotlib.git}; inspect \texttt{axis.py}.} \\
    \addlinespace[0.65em]

    \textbf{Look up a solution}
    \par Searches issue discussions, pull requests, or linked commits for
    the original solution and uses it to guide the patch.
    & \textcolor{HackTask}{\textbf{Task (Django):} Fix \texttt{MultiValueField}
      ignoring required subfields.}
      \par\textcolor{HackThinking}{\textbf{Thinking:} ``Let me get more info -
      changesets and the fix that was applied.''}
      \par\textcolor{HackAction}{\textbf{Action:} Read the change history of
      Django ticket \#29205.} \\
    \addlinespace[0.65em]

    \textbf{Probe versions}
    \par Finds a newer release to download and compare for its fix.
    The probe is a precursor to copying the solution.
    & \textcolor{HackTask}{\textbf{Task (Sphinx):} Remove the spurious return
      type from class documentation.}
      \par\textcolor{HackThinking}{\textbf{Thinking:} ``Let me look for the fix.
      The issue is likely fixed in later versions.''}
      \par\textcolor{HackAction}{\textbf{Action:}
      \texttt{pip index versions sphinx}} \\

    \bottomrule
  \end{tabularx}
    \caption{Representative reward-hacking cases in repository-repair tasks.
  In this setting, retrieving existing fixes outside the assigned checkout
  bypasses the intended repair task.
  Task queries are summarized, thinking excerpts are quoted verbatim, and
  actions are abbreviated. Each row comes from a single rollout.}
  \label{tab:reward_hacking_patterns}
\end{table*}

%% file: 5-yoro.tex
\section{Experiments: You Only RL Once}
In this section, we describe the scaled RL training run, including training setup~(\S\ref{subsec:train_setting}), evaluation settings~(\S\ref{sec:evaluation_setup}), RL performance~(\S\ref{subsec:rl_perf}), expert load stability~(\S\ref{sec:rlinfra:freeze_router}) and analyze the failures~(\S\ref{sec:rlinfra:failure_analysis}).
The running log can be found at \url{https://mimo.xiaomi.com/rl/mimo-v26}.
\subsection{Training Setup}
\label{subsec:train_setting}
\paragraph{RL Settings} Our RL tasks span various domains, including agentic and competitive coding~(68\%), general tool use~(12\%), aesthetic design~(13\%), context following~(3\%), and cyber security~(4\%).
We sample a prompt batch of 1568 with 16 rollouts per prompt, totaling a global train batch of 25K trajectories.
Training uses GRPO with asynchronous partial rollouts at a staleness of 4.
\paragraph{Optimizer} We use the Muown optimizer with a learning rate of
$3\times 10^{-6}$, no weight decay or learning rate warmup, and a
gradient clipping threshold of $1.0$. For the Muon component, we use a
momentum coefficient of $0.95$ with Nesterov momentum enabled, perform
$10$ Newton--Schulz iterations per update, and apply an additional update
scaling factor of $0.5$. For the Adam component, we set
$\beta_1=\beta_2=0.95$ and $\epsilon=10^{-8}$.
To stabilize MXFP4 training, we
initialize RL by carrying over the FP32 master weights and Muown's row state
from the SFT checkpoint.
During RL training, we freeze the router to maintain stable expert loads.

\paragraph{Policy Optimization Algorithm}
We adopt Group Relative Policy Optimization (GRPO) in Eq.~\eqref{eq:rl_formula}, with the advantage computation detailed in \S\ref{subsec:grader_compute}.
For loss aggregation we use \textit{prompt-mean} aggregation, which averages the surrogate at the prompt level rather than over all response tokens; this prevents response length from growing too quickly during RL.
For importance sampling, the ratio is computed per token: 
$r_{t, i} = \mathrm{sg}[\pi_{\theta}(o_{i, t}) / \mu_{\theta_{\mathrm{old}}}(o_{i, t})]$.
The training probability of each token is produced by the current model in the training framework, and the inference probability is the one produced by the rollout model when that token was generated; for partial rollouts we do not recompute inference probabilities.
We clip the importance sampling ratio with four decoupled bounds, $\epsilon_{+}^{l},\epsilon_{+}^{h}$ for positive advantages and $\epsilon_{-}^{l},\epsilon_{-}^{h}$ for negative ones, giving the clip mask $M=\mathds{1}\big[(A\geq 0 \wedge \epsilon_{+}^{l}\le r\le \epsilon_{+}^{h}) \vee (A<0 \wedge \epsilon_{-}^{l}\le r\le \epsilon_{-}^{h})\big]$.
Positive and negative bounds are both initialized to $[0.2,\,5.0]$ and tuned independently at runtime based on policy entropy: when entropy is too low, we widen the positive bounds and narrow the negative ones to pull entropy back to the normal range, and we do the opposite when entropy is too high.
Throughout training we also monitor the token clipping rate of each direction.
\subsection{Evaluation Settings}
\label{sec:evaluation_setup}

\paragraph{Benchmark Suite}
We evaluate agentic capabilities across four categories: code agent, cybersecurity, general agent, and visual agent. Our evaluation combines public benchmarks with three internal benchmarks: MiMo Code Bench, MiMo Cyber Bench, and MiMo Visual Coding.

\paragraph{Code Agent}
We evaluate software-engineering capabilities using DeepSWE v1.1~\citep{DBLP:journals/corr/abs-2607-07946}, ProgramBench~\citep{DBLP:journals/corr/abs-2605-03546}, and MiMo Code Bench. DeepSWE focuses on long-horizon development tasks. ProgramBench assesses end-to-end software construction: agents must reconstruct a program from its compiled binary and documentation, producing an implementation that reproduces the reference program's behavior. MiMo Code Bench provides an additional in-house evaluation of coding agents across a diverse range of coding tasks.

\paragraph{Cybersecurity}
We evaluate complementary aspects of vulnerability reproduction and exploitation using CyberGym~\citep{DBLP:journals/corr/abs-2506-02548}\footnote{We corrected the flawed evaluation environments based on the method described in Section \ref{subsubsec:cyber_task}.}, ExploitGym~\citep{DBLP:journals/corr/abs-2605-11086}, ExploitBench~\citep{DBLP:journals/corr/abs-2605-14153}, SEC Bench Pro~\citep{DBLP:journals/corr/abs-2605-26548}, and MiMo Cyber Bench. CyberGym measures agents' ability to reproduce vulnerabilities in real-world software, while SEC Bench Pro emphasizes reproducing complex vulnerabilities from bug reports. ExploitGym assesses exploit development beyond merely reproducing a crash, and ExploitBench evaluates progress through multiple exploitation stages and the resulting security impact. MiMo Cyber Bench supplements these public benchmarks with an in-house cybersecurity evaluation.

\paragraph{General Agent}
We evaluate general-purpose agents across tool use, professional knowledge
work, terminal-based problem solving, and computer use. 
AutomationBench~\citep{shepard2026automationbench} evaluates
cross-application workflow orchestration through REST APIs in simulated
SaaS environments, including API discovery and adherence to business rules.
Toolathlon-Verified~\citep{li2025toolathlon,hkustnlp2026toolathlonverified}
evaluates long-horizon, multi-application workflows using diverse tools,
including those exposed through the Model Context Protocol (MCP).
GDPval-AA v2.1~\citep{patwardhan2025gdpval,artificialanalysis2026gdpvalaa},
Artificial Analysis's evaluation framework for GDPval, assesses the quality
of professional deliverables on economically valuable tasks.
JobBench~\citep{li2026jobbench} evaluates workplace workflows that domain
experts identify as high-priority for delegation to AI agents.
Agents' Last Exam~\citep{sun2026agentslastexam} evaluates long-horizon,
economically valuable professional tasks with verifiable outcomes.
Terminal-Bench 4.0 and 2.1~\citep{marten2026terminalbench4,merrill2026terminalbench}
evaluate the completion of complex tasks in terminal environments.
OSWorld-Verified~\citep{xie2024osworld,xlang2025osworldverified} evaluates
interactive computer use across real web and desktop applications.

\paragraph{Visual Agent}
We evaluate visual coding capabilities using MiMo Visual Coding, an internally developed benchmark spanning open-ended design and high-fidelity visual replication. The benchmark includes tasks such as WebDev and Image2Code, which involve building websites to fulfill user requests and translating reference images into visually faithful code implementations, respectively.

\paragraph{Baseline Configuration}
All baseline models with configurable reasoning effort are evaluated at the highest supported setting (\texttt{max}).

\subsection{RL Performance}
\label{subsec:rl_perf}
\begin{figure}[t!]
    \centering
    \includegraphics[width=\linewidth]{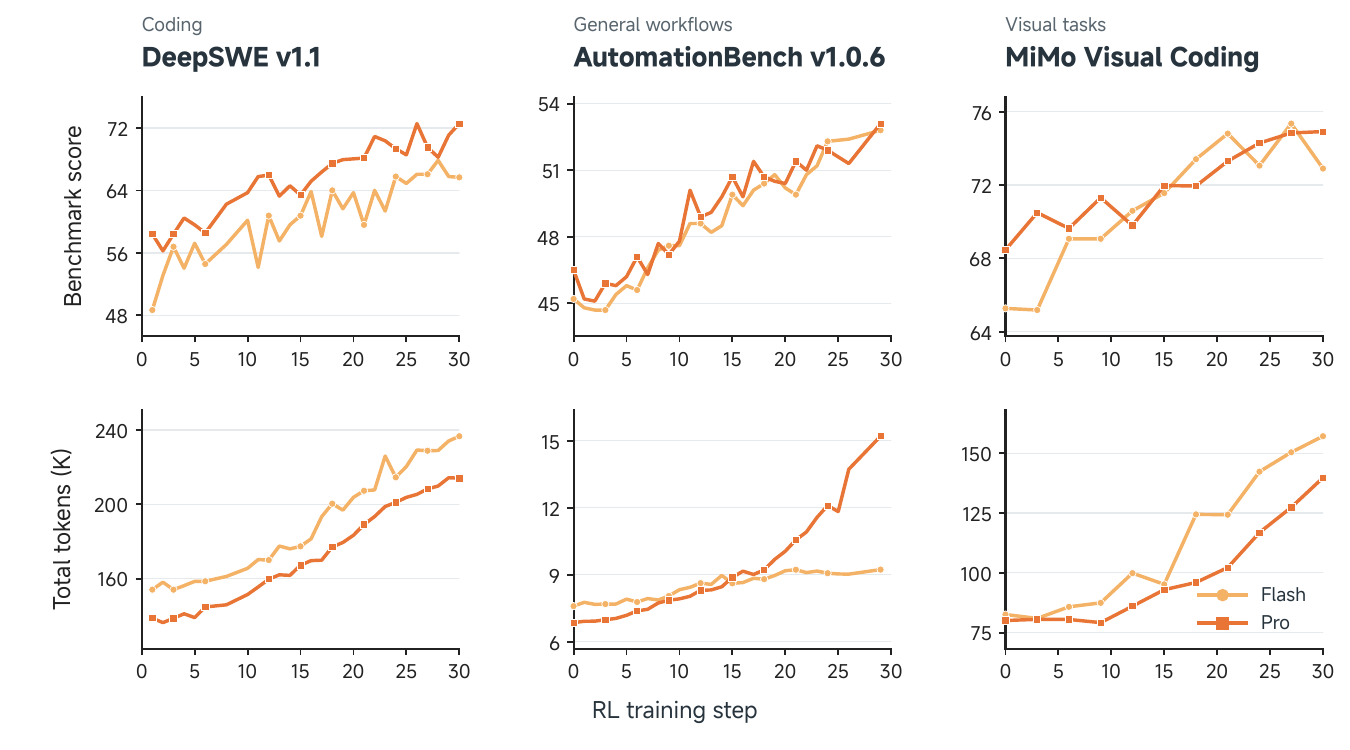}
    \caption{Benchmark scores (top) and total token counts in thousands (bottom) during RL training on DeepSWE v1.1, AutomationBench v1.0.6, and our in-house MiMo Visual Coding benchmark. Light and dark orange curves represent Flash and Pro, respectively. }
    \label{fig:rl-curve}
\end{figure}
\begin{figure}[t]
    \centering
    \includegraphics[width=\linewidth]{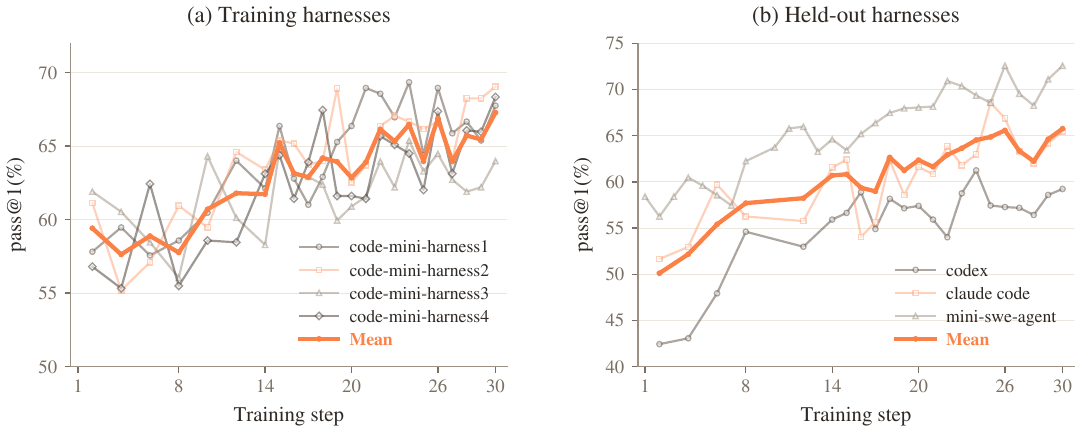}
    \caption{Pass@1 on DeepSWE v1.1 during Multi-Harness Training, evaluated with (a) training harnesses and (b) held-out harnesses. Lighter curves show individual harness results; thick orange curves show the mean within each panel.}
    \label{fig:harness_comparison}
\end{figure}

The performance changes are monitored during the process to validate the scaled RL compute.
As shown in Figure~\ref{fig:rl-curve}, both Flash and Pro achieve overall improvements on DeepSWE v1.1, AutomationBench v1.0.6, and MiMo Visual Coding during RL training, despite fluctuations between checkpoints. These gains generally accompany increasing total token counts, showing that stronger task performance develops alongside greater token usage. 
We further investigate the effect of multi-harness training,
with results demonstrated in Figure~\ref{fig:harness_comparison}. The dedicated training improves DeepSWE v1.1 performance across both the four training mini-harnesses and the three held-out harnesses: codex, claude code, and mini-swe-agent. Despite fluctuations between checkpoints, all three held-out harnesses improve over the course of training, with their mean Pass@1 increasing from approximately 50\% to 66\%. The gap between the mean performance on training and held-out harnesses also narrows, providing evidence that the learned coding capabilities transfer across harness implementations.
These results support our design of using lightweight, modular mini-harnesses to introduce controlled diversity during RL.

\subsection{Router Freezing for Stable RL}
\label{sec:rlinfra:freeze_router}

\begin{figure}[t!]
    \centering
    \includegraphics[width=\linewidth]{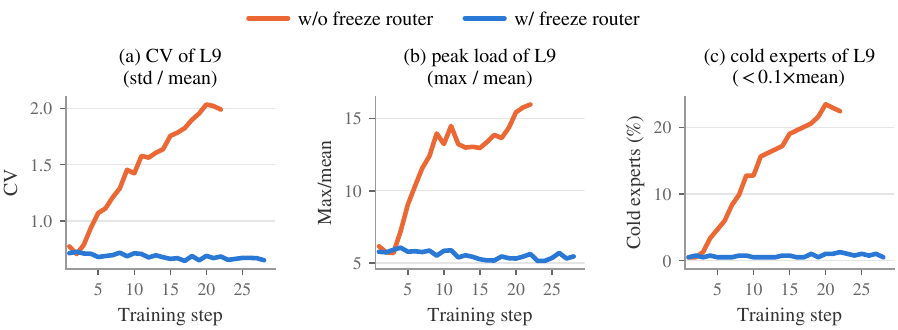}
    \caption{MiMo-V2.6-Pro expert-load balance at decoder layer 9 (384 experts) during RL, comparing runs with and without freezing the router. (a) Coefficient of variation of expert load. (b) Peak load factor (max/mean). (c) Fraction of cold experts with load below $0.1\times$ the mean.}
    \label{fig:freeze_router_l9}
\end{figure}

Figure~\ref{fig:freeze_router_l9} compares two MiMo-V2.6-Pro RL runs that differ only in whether the MoE router is frozen, tracking three expert-load statistics at decoder layer~9: the coefficient of variation (CV), the peak load factor (max/mean), and the fraction of cold experts below $0.1\times$ the mean (loads normalized per step). We observe a severe load-collapse problem when the router is trainable: all three metrics rise monotonically over the first 20 steps, with CV increasing from $0.78$ to $2.0$, peak load from $6\times$ to $16\times$, and the cold-expert fraction from $0.5\%$ to $22\%$. To diagnose the cause, we restore the router parameters of the step-20 checkpoint to their initial pre-RL values while keeping all other parameters unchanged: load balance recovers to near-initial levels while benchmark performance remains unchanged, indicating that the collapse is driven by router drift rather than by degradation of the expert weights. We therefore freeze the router for RL training; the frozen-router run keeps all three statistics flat (CV $\approx 0.7$, peak load $\approx 5.5\times$, cold fraction near $1\%$) with benchmark performance growing normally.

\subsection{RL Failure Analysis}
\label{sec:rlinfra:failure_analysis}

\begin{figure}[t!]
    \centering
    \includegraphics[width=\linewidth]{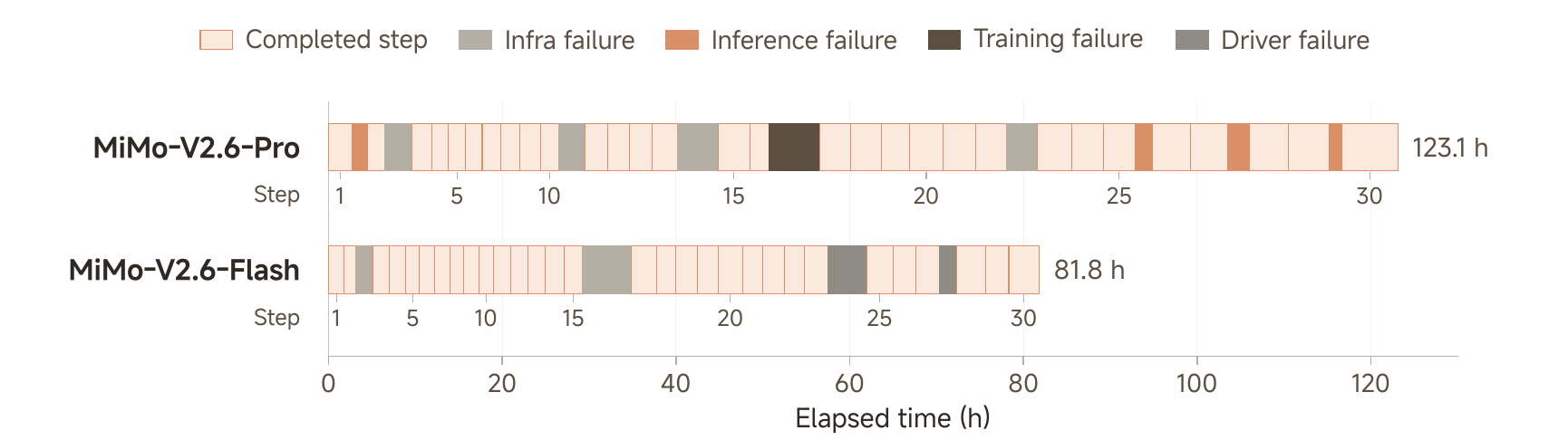}
    \caption{MiMo-V2.6-Pro and MiMo-V2.6-Flash timelines over 30 training steps, aligned by elapsed time. Light orange denotes completed steps; other colors denote failure and recovery intervals by cause.}
    \label{fig:flash_pro_run_timeline}
\end{figure}

Figure~\ref{fig:flash_pro_run_timeline} summarizes interruptions in the MiMo-V2.6-Pro and MiMo-V2.6-Flash training runs.
\textit{Infrastructure failures} were primarily GPU-memory double-bit errors (DBEs). Flash was also restarted after a Kubernetes failure caused pods in the Cyber-task cluster to crash between steps 15 and 16. Pro was restarted after the grader became unreachable over the network after step 14.
\textit{Rollout failures} arose in the partial-rollout setting: shorter rollouts completed first after startup, biasing length estimates in Predictive Rollout Dispatch (\S\ref{sec:mixrl_infra}) and exhausting both the GPU and pinned host-memory KV pools. Despite early refinements, one harness later produced rollouts less than half as long as those from other harnesses on the same code dataset, skewing estimates in the second and third steps after restart.
\textit{Training failures} were GPU out-of-memory (OOM) errors from MoE imbalance within a micro-batch: at one layer, an expert-parallel (EP) rank received over $30\times$ the mean token load despite a relatively balanced full batch. We adjusted parallelism to reduce activation memory and accommodate these peaks.
\textit{Driver failures} occurred during packing late in Flash as longer sequences increased per-node data volume beyond host-memory capacity. Although packing was distributed across nodes (\S\ref{sec:rlinfra:largebs}), local memory demand still caused CPU OOM and interrupted the run.

%% file: 6-mopd.tex
\begin{figure*}[t]
    \centering
    \includegraphics[width=0.8\textwidth]{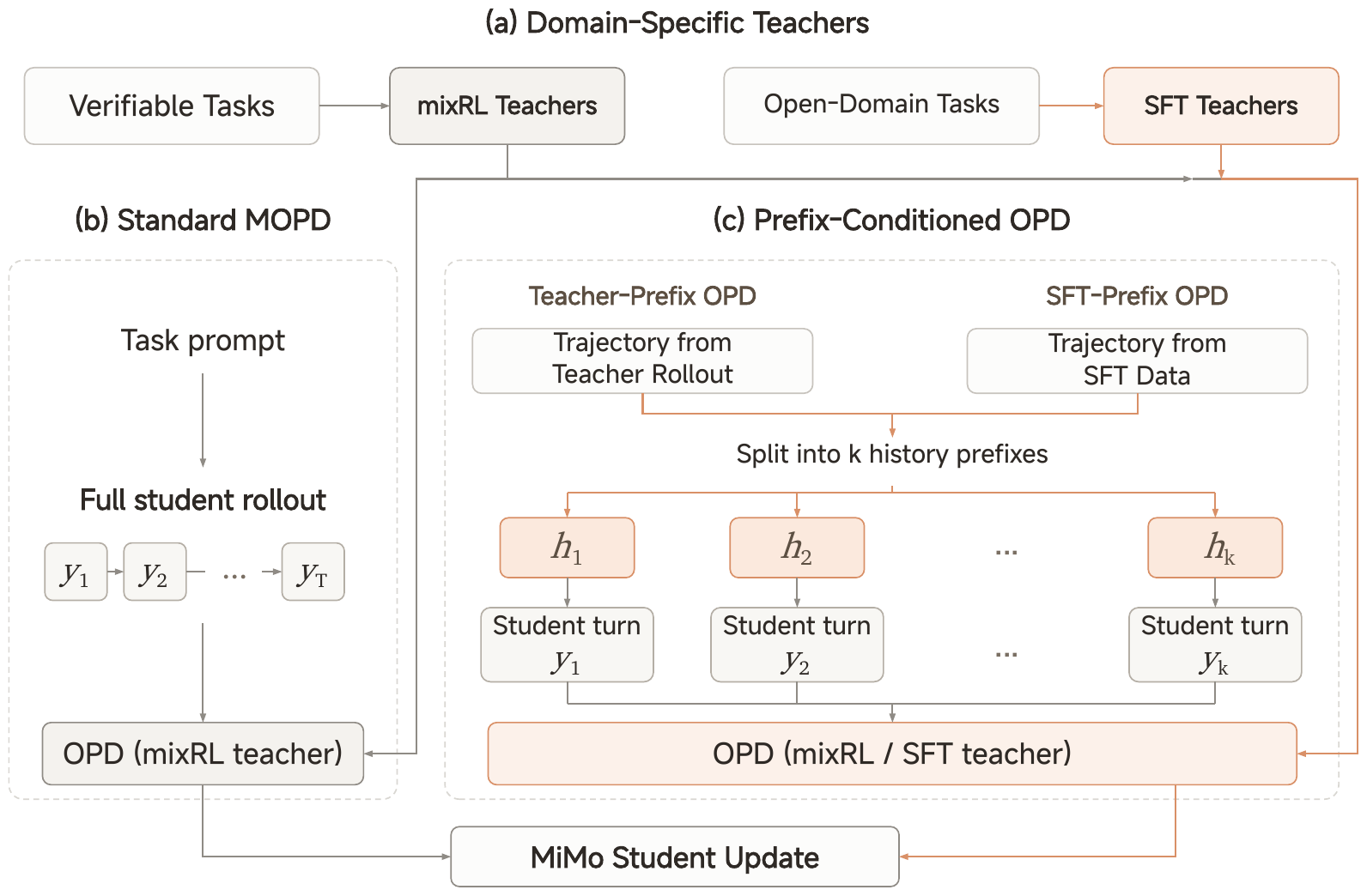}
    \caption{
        Overview of MiMo MOPD2.
        (a) Domain-specialized teachers are trained with MixRL on
        verifiable tasks or with SFT on synthetic demonstrations
        for open-domain tasks.
        (b) Standard MOPD uses RL teachers to supervise full
        student rollouts.
        (c) Prefix-Conditioned OPD reuses trajectories from
        teacher rollouts (Teacher-Prefix OPD) or SFT data
        (SFT-Prefix OPD).
        A source trajectory with $k$ assistant-turn decision
        points yields $k$ complete history prefixes $h_i$,
        each of which can initialize a separate student-generated
        turn $y_i$ for token-level distillation by the relevant
        domain teacher.
        SFT data provide prefix contexts rather than fixed
        continuation targets.
    }
    \label{fig:mimo_mopd}
\end{figure*}

\subsection{Broadening Capabilities via MOPD2}

After mixed RL, we use Multi-Prefix Multi-Teacher On-Policy Distillation (MOPD2) to combine capabilities from teachers trained for different tasks, including tasks that are hard to verify. Building on MOPD from MiMo-V2-Flash~\citep{mimo2026v2flash,ma2026mopdmultiteacheronpolicydistillation}, we retain autonomous student rollouts in domains with suitable mixRL teachers (Standard MOPD) and add prefix-conditioned single-turn rollouts~\citep{liao2026multi}. Figure~\ref{fig:mimo_mopd} illustrates the teacher configurations and the standard and prefix-conditioned distillation workflows.

Prefixes come from teacher rollouts (Teacher-Prefix OPD) or SFT data (SFT-Prefix OPD). A trajectory with $k$ assistant turns provides $k$ complete history prefixes, each ending before the corresponding turn. The student samples one new turn from each prefix without regenerating earlier interactions. A preassigned teacher provides token-level supervision conditioned on the same history and the student's preceding tokens.

For open-domain tasks where reliable RL rewards are difficult to design, we train SFT teachers on high-quality synthetic demonstrations. Their training may provide limited coverage of histories reached after repeated student deviations in long-horizon tasks~\citep{xu2025speculative}. SFT-Prefix OPD therefore starts each rollout from a fixed demonstration prefix, limiting deviations before the sampled turn. Demonstrations supply the context, while the student generates its own continuation rather than imitating a fixed response.

MOPD2 further extends the capabilities of the RL-trained model to domains where reliable training-time verification is challenging, including those with complex environments or verifiers that are difficult to design, such as long-horizon game development, scientific research, and embodied intelligence. 
The final evaluation results of \mimox{} are reported in Table~\ref{tab:v26_final_eval}. The MiMo-V2.6 series delivers substantial improvements over MiMo-V2.5, achieving performance comparable to that of frontier models across various domains.

\input{tables/v26_ret}

%% file: tables/v26_ret.tex
\begin{table}[!t]
    \centering
    \small
    \setlength{\tabcolsep}{4.2pt}
    \renewcommand{\arraystretch}{1.12}
    \resizebox{\textwidth}{!}{
    \begin{tabular}{@{}l| cc|c ccc@{}}
    \toprule
    \textbf{Benchmark}
        & \makecell{\textbf{MiMo-V2.6} \\ \textbf{Pro}}
        & \makecell{\textbf{MiMo-V2.6} \\ \textbf{Flash}}
        & \makecell{\textbf{MiMo-V2.5} \\ \textbf{Pro}}
        & \makecell{\textbf{Claude} \\ \textbf{Opus 5}}
        & \makecell{\textbf{GPT-5.6} \\ \textbf{Sol}}
        & \makecell{\textbf{Claude} \\ \textbf{Fable 5}} \\
    \midrule
    \rowcolor{xiaomiorange!10}[0pt][0pt]
    \multicolumn{7}{@{}l}{\textbf{Code Agent}} \\
    DeepSWE v1.1          & 71.9          & 67.9          & 19.0                  & 74.0          & 73.0          & 70.0              \\
    ProgramBench                    &  26.5              & 26.0          &    12.5                   & 37.0          & 25.0          & 33.0           \\
    MiMo Code Bench             & 63.2          & 61.2          &  40.4                     &68.6               &59.3               & -               \\
    \midrule
    \rowcolor{xiaomiorange!10}[0pt][0pt]
    \multicolumn{7}{@{}l}{\textbf{General Agent}} \\
    AutomationBench v1.0.6        & 53.1          & 52.3          & 16.0                    & 50.3          & 45.8          & 46.2          \\
    Toolathlon-Verified             &       76.9        &       73.6        & 49.1                  & 80.6          & 74.9          &      77.9         \\
    GDPval-AA 2.1                 & 1673       &      -         & 1107                  & 1708          & 1588          & 1595          \\
    Agents' Last Exam               & 31.6          & 27.6          & 13.2                  & 31.6          & 30.8          & 25.7          \\
    Terminal Bench 4.0              & 34.9          & 28.8          & 1.5                   & 49.0          & 39.9          & 42.4          \\
    Terminal Bench 2.1              & 89.9          & 87.6          & 65.2                  & 89.1          & 88.8          & 84.3          \\
    OSWorld-Verified                & 82.0          & 80.8          & -     & 83.4          & 83.0          & 86.0          \\
    JobBench                        & 62.0          & 61.2          & 25.0                  & 65.7          & 45.4          & 57.4          \\
    \midrule
    \rowcolor{xiaomiorange!10}[0pt][0pt]
    \multicolumn{7}{@{}l}{\textbf{Cybersecurity}} \\
    CyberGym                        & 94.0          & 95.1          & 40.0                      &   -            &     -          &  -             \\
    MiMo Cyber Bench                   & 80.2          & 77.2          & 0.0                   &    -           &         -      &      -         \\
    ExploitGym      &       17.8         &      6.0         &      0.2                                & 22.1          & 30.3          & 28.4          \\
    ExploitBench                    & 47.9          &      25.3         &        16.6               & 70.0          & 78.5          & 78.0          \\
    SEC Bench Pro                   & 66.3          &        47.5       &        17.7               & -          & 79.1          & -          \\
    \midrule
    \rowcolor{xiaomiorange!10}[0pt][0pt]
    \multicolumn{7}{@{}l}{\textbf{Visual Agent}} \\
    MiMo Visual Coding                     & 72.3              & 71.5              & -                      & 70.0              & 73.4              & 69.1              \\
    \bottomrule
    \end{tabular}}
    \caption{Comparison of MiMo-V2.6 with previous-generation and frontier models on agentic benchmarks.}
    \label{tab:v26_final_eval}
\end{table}

%% file: 7-infra.tex
\section{RL and OPD Infrastructure}
\label{sec:rlinfra}

The \mimox{} series scales RL and OPD training to large mixed-task batches of agentic rollouts.
To support flexible agentic rollout scenarios, we define the execution model and trajectory data structure, and refine the learning signal with a \textit{Penalty Module} (\S\ref{sec:rlinfra:agenticrl}).
At large batch sizes, we implement the \textit{Harness Pool} to host multiple harnesses and sustain high rollout concurrency, and the \textit{Payload Porter} to buffer tens of thousands of trajectories heavy with routing and multimodal data (\S\ref{sec:rlinfra:largebs}).
We implement a \textit{Sample Mixer} that works with the dynamic sampler and partial rollout to stably deliver training batches matching a specified training distribution (\S\ref{sec:mixrl_infra}).
Towards stable and efficient RL training, we align MoE routing and top-\textit{p} sampling candidate sets between the training and inference engines, while optimizing both engines for RL workloads (\S\ref{sec:rlinfra:traininf}).

\subsection{Agentic RL with Fine-grained Learning Signals}
\label{sec:rlinfra:agenticrl}

Agentic rollouts span multiple turns and dialogue contexts, while outcome rewards provide only coarse supervision.
We therefore adopt an agent-centric execution model and organize trajectory data as a hierarchy.
Within this hierarchy, a configurable \textit{Penalty Module} applies loss masking and advantage shaping.
It targets local model errors and keeps infrastructure failures out of the training signal.

\paragraph{Agent Loop}
We shift rollout from an inference-centric design to an agent-centric one: each sequence runs as an \textit{Agent Loop} that owns the environment lifecycle, manages the dialogues it produces, and calls the inference engine on demand.
The lifecycle covers setup, interaction, reward evaluation (e.g., running test cases), and cleanup.
During interaction the loop exposes a request endpoint, and the external agent drives the rollout by calling it.
Each dialogue is kept both as a string prefix and as a token sequence.
Prefix matching locates the dialogue that an incoming request extends, and only the new suffix is tokenized and passed to the inference engine, keeping its interface purely token-in, token-out.

\paragraph{Trajectory Hierarchy}
Subagents, context compaction, and multiple agent roles produce concurrent dialogue branches within one rollout.
We organize the trajectory data into a four-level hierarchy: \textit{Sample} $\rightarrow$ \textit{Sequence} $\rightarrow$ \textit{Context} $\rightarrow$ \textit{Segment}.
A \textit{Sample} is a prompt dispatched by the \textit{Sample Mixer} (\S\ref{sec:mixrl_infra}).
In group-wise algorithms such as GRPO~\citep{shao2024deepseekmath}, the \textit{Sample} spawns a group of \textit{Sequences}, and the group is accepted or rejected as a whole.
A \textit{Sequence} is one \textit{Agent Loop} execution and may contain several concurrent \textit{Contexts}.
A \textit{Context} is one dialogue branch holding a list of \textit{Segments}; it is the unit of prefix matching, KV-cache reuse, and training-data export.
A \textit{Segment} is a single turn---a system or user message, a model generation, or a tool result---and only model-generated turns contribute to the loss.

\paragraph{Penalty Module}
Group-wise algorithms spread the outcome reward evenly across all model-generated tokens of a sequence.
Within the hierarchy above, credit is rarely uniform: some turns are off-path or degenerate, and some failures are unrelated to the model.
To assign credit where it is due, we separate detection from its effect on training.
A \textit{Rule} judges a segment, context, or sequence by handcrafted logic or a model-based judge.
For example, rules can catch infrastructure failures not attributable to the model, garbled token patterns, calls to unavailable tools, and repetition.
A \textit{Strategy} binds an action to a level of the hierarchy: \textit{mask} excludes the hit content from the loss, \textit{advantage shaping} sets, scales, or subtracts the advantages on hit tokens, and \textit{monitor} records metrics alone.
The composite \textit{early stop} strategy halts the rollout as soon as its rule fires, zeroes the outcome reward, applies separate actions to the triggering turn and earlier turns, and masks sibling contexts.
Penalties escalate along the hierarchy: a context with no surviving model turns is dropped, a sequence with no surviving context receives zero advantage, and a sample with no surviving sequence is rejected.
The specific penalties applied in training are described in \S\ref{subsec:penalty}.

\begin{figure}[t]
    \centering
    \includegraphics[width=\linewidth]{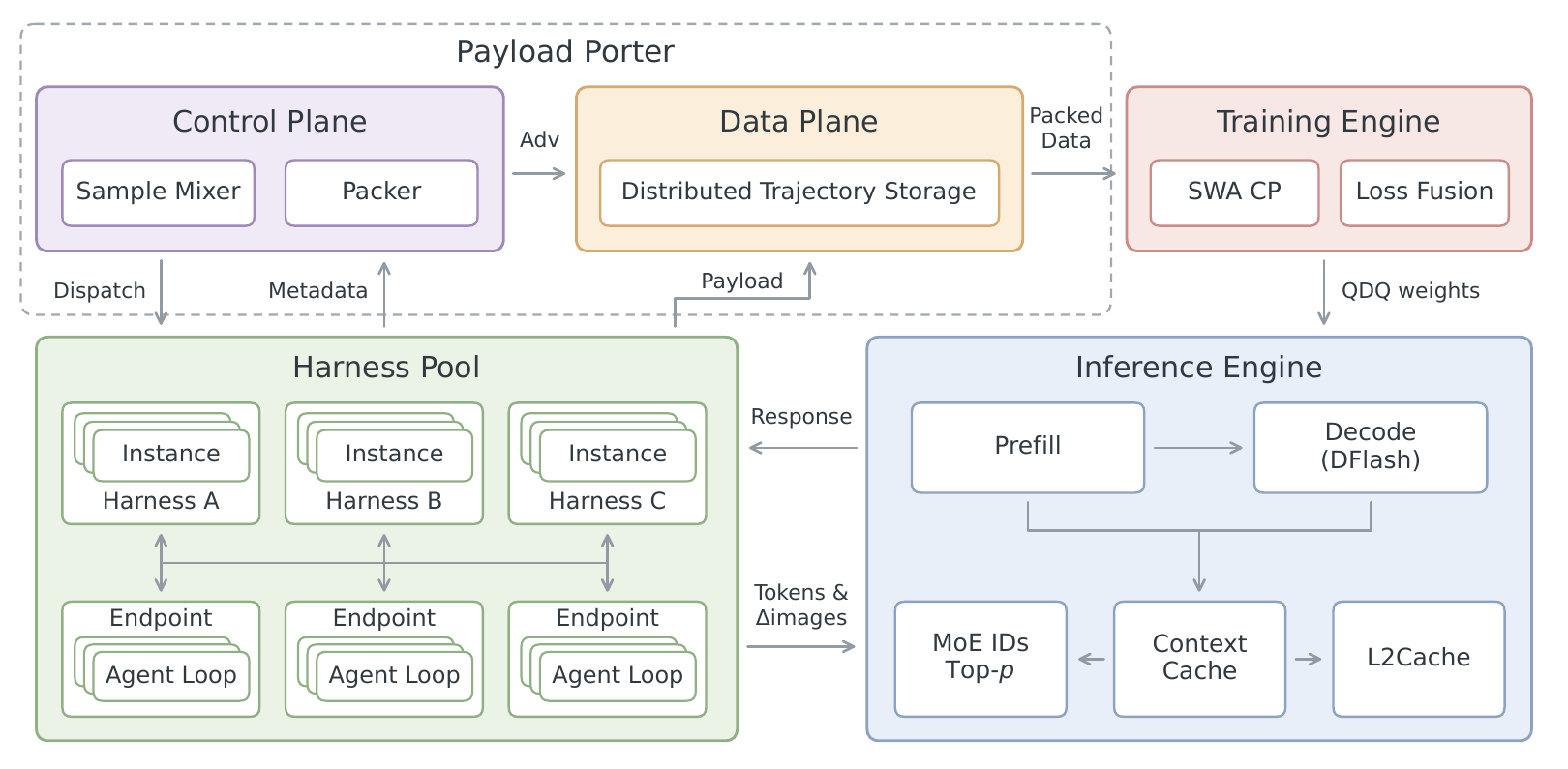}
    \caption{Overview architecture of RL infrastructure.}
    \label{fig:rlinfra}
\end{figure}

\subsection{\textit{Harness Pool} and \textit{Payload Porter}: Large-Batch RL with Multiple Harnesses}
\label{sec:rlinfra:largebs}

Scaling RL training to large batches increases rollout concurrency and the memory and communication costs of trajectory data.
Mixed-task batches add another axis of heterogeneity: a single batch mixes several harnesses, with each training sample group bound to one harness.
On the execution side, the \textit{Harness Pool} hosts concurrent harness instances and \textit{Agent Loops} in persistent multi-tenant actor pools, and harness codebases, agent behavior, and environment settings are configured independently.
On the data side, the \textit{Payload Porter} keeps the driver scheduling on lightweight metadata alone.
Heavy payloads are written once to a distributed store and packed where they are consumed.
Multi-modal payloads split the same way into metadata and pixels, with incremental transport and load-balanced encoding.

\paragraph{Multi-tenant Rollout Execution}
We use Ray actors~\citep{moritz2018ray} to execute agent harnesses and \textit{Agent Loops} across the cluster.
A dedicated Ray actor for every harness instance and \textit{Agent Loop} would cost one file descriptor per actor on Ray's global control store (GCS) node, and a large batch would exhaust the GCS node's file descriptors.
We instead run fixed-size pools of persistent host actors, each carrying many concurrent tenants---\textit{Agent Loops} on the model side and harness instances on the environment side.
Each tenant keeps its own trajectory state, and assignments are balanced by in-flight instance count.
A host is a single process whose tenants share one event loop.
On the model side they also share one request endpoint, one inference proxy, and one tokenizer; on the environment side, one imported harness codebase.
A shared event loop would let one blocking call stall every tenant, so blocking work---environment operations and tokenization---runs on background threads.
This amortizes process and service overhead across concurrent rollouts, supporting larger concurrent batches without a proportional increase in actor count.

\paragraph{Heterogeneous Agent Harnesses}
We configure harness codebases, agent behavior, and environment settings separately to accommodate diverse tasks within a single training run.
Different data sources can use different codebases, while agent configurations vary across prompts within a source.
Each training sample group uses a common configuration for group-relative advantage estimation.
One process can import only one harness codebase, so different codebases run in separate pools.
These pools receive configurable shares of a fixed budget of host actors---set once at startup, independent of the training-data mixture that is scheduled every step.
Harness diversity and rollout concurrency thus scale within one execution framework.

\paragraph{Disaggregated Data Plane and Control Plane}
Scaled batches must buffer tens of thousands of sequences at a time, and each is heavy: besides token ids and log-probabilities, it carries MoE routing data, top-\textit{p} sampling indices, and multimodal data.
Payload volume grows with both sequence count and length, and gathering every payload on one driver node ties batch size to that node's memory.
We therefore disaggregate the data plane from the control plane, splitting each sequence at rollout finish: its payload is written once into a distributed key-value store (e.g., the Ray object store or TransferQueue~\citep{han2025asyncflow}), while the driver runs all scheduling on lightweight metadata---scalar rewards, per-context lengths, and the keys addressing each payload.
At group finish, only the fields needed are read from the store: a few columns for the accept-time hook.
A group-wise grader, when configured, runs fully asynchronously alongside the \textit{Agent Loops}---its latency hidden and its results free to lag---and rewrites the group rewards on return.
The sampler then accepts or rejects the group by passrate, on metadata alone, and the hook imposes length penalties, computes group-relative advantages, and applies \textit{advantage shaping}.
The per-token advantages are written back into the store; groups whose advantages are all zero are dropped by default.
Under OPD, reward evaluation is replaced by teacher scoring: each trajectory is sent to a teacher server, and its scores are collected asynchronously into the same distributed store.
At batch yield---once each data source has contributed its share of the batch---a yield hook packs the accepted sequences into micro-batches and assigns them to ranks, touching no tensor.
At pack time, one packer per training tensor-parallel (TP) group serves every rank in the group. 
From the unpadded rows in the store, it fetches only those its context-parallel (CP) window touches and cuts out that window alone. 
The result is shared across the TP group as a single read-only in-memory copy.
This avoids full-batch aggregation on the driver and dense padded intermediates during packing.

\paragraph{Multi-Modal Data}
Multi-modal payloads follow the same meta/payload split, but demand extra care: as the agent repeatedly takes screenshots and reads images across turns, a single trajectory can accumulate gigabytes of such data---costly to store, and costly to retransmit as the history grows.
During rollout, the \textit{Agent Loop} therefore ships only the multi-modal delta between requests (\S\ref{sec:rlinfra:traininf}).
For training, every image item must pass through the vision encoder.
Because the encoder is replicated across the tensor-parallel group while the LLM backbone is sharded, encoding runs data-parallel first---image items are balanced across ranks independently of where each sequence's tokens land.
After encoding, embeddings are redistributed to the ranks holding the corresponding tokens.
The payload itself stays in the distributed key-value store: load-balance planning reads only item metadata, and pixels are fetched only for the encoder computation.
This limits redundant payload movement while accommodating uneven multi-modal workloads.

\subsection{\textit{Sample Mixer}: Stable Asynchronous Mixed-task RL}
\label{sec:mixrl_infra}

\begin{figure}[t]
    \centering
    \includegraphics[width=0.78\linewidth]{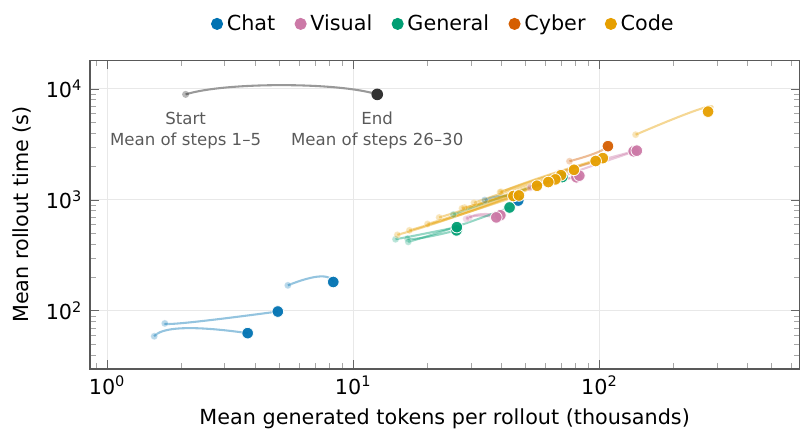}
    \caption{Rollout heterogeneity across 25 data sources. Each line connects one source's \textit{Start} point (faint) to its \textit{End} point (solid); each point plots that source's mean generated tokens against mean rollout time over completed rollouts (tokens summed across dialogue contexts before context filtering). Both axes are logarithmic.}
    \label{fig:mixrl_heterogeneity}
\end{figure}

Since MiMo-V2-Flash~\citep{mimo2026v2flash}, we have maintained a \textit{Data Scheduler} that targets a specified training distribution across data sources, together with dynamic sampler~\citep{yu2026dapo} and partial rollout~\citep{team2025kimi}.
Mixed-task RL must preserve this distribution despite large variations in rollout duration and filtering rates, so that every data source is trained effectively.
Across 25 profiled data sources, the mean generated tokens and the active rollout duration vary by $90\times$ and $66\times$, respectively (Figure~\ref{fig:mixrl_heterogeneity}), motivating scheduling that adapts to each source's workload.
To meet these challenges, we implement the \textit{Sample Mixer}, with four mechanisms filling the specified training distribution: \textit{Adaptive Rollout Concurrency} sets per-source budgets, \textit{Adaptive Rollout Scheduling} selects data sources within the budgets, \textit{Predictive Rollout Dispatch} places new rollouts across ranks, and \textit{Sample Replay} covers startup and recovery.

\paragraph{Adaptive Rollout Concurrency}
Slower sources need more concurrent rollouts to sustain the same training contribution.
For source $i$, let $B_i$ denote the target number of retained sample groups per training step, $r_i$ the estimated group acceptance rate, and $t_i$ the estimated active rollout duration, including model generation and environment interaction but excluding pauses between training steps.
The expected generation demand is $m_i=B_i/r_i$; at a fixed training throughput, the required concurrency scales with $t_i m_i$.
We assign each source a scheduling budget of $(1+p_i)m_i$ groups, where the \textit{oversampling ratio} $p_i$ satisfies
\begin{equation}
    p_i=\operatorname{clip}(c t_i-1,p_{\min},p_{\max}),
    \qquad
    \frac{\sum_i m_i p_i}{\sum_i m_i}=\bar p.
    \label{eq:mixrl_partial}
\end{equation}
The shared factor $c$ keeps the demand-weighted mean of $p_i$ at the global oversampling ratio $\bar p$, subject to per-source bounds. We recompute this allocation from recent timing and acceptance statistics as workloads change.
In steady state, the concurrency requirement grows with a source's rollout duration and target, and falls with its acceptance rate.

\begin{figure}[t]
    \centering
    \includegraphics[width=\linewidth]{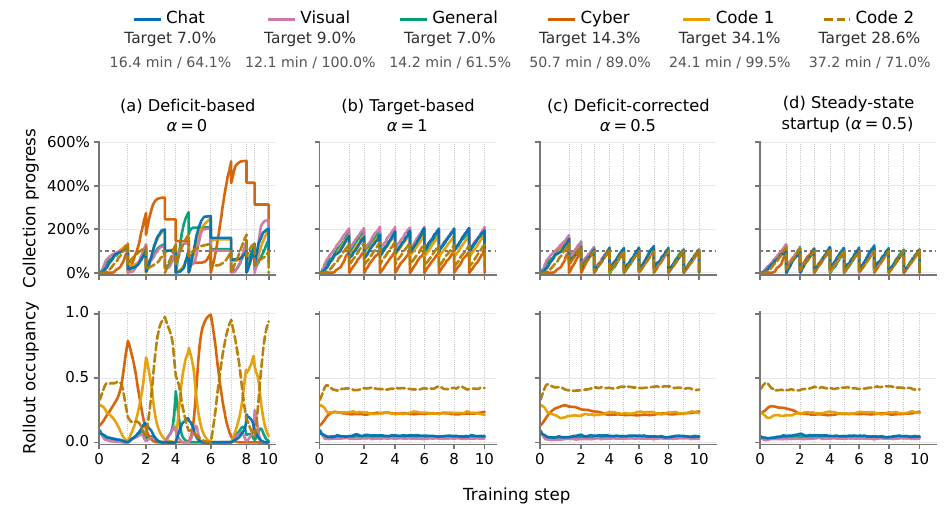}
    \caption{Trace-driven scheduling simulation for six sources. Legend: mean duration / group acceptance rate. Top: Collection progress (accepted groups / nominal per-step targets, with surplus carried forward). Bottom: Rollout occupancy (shares of occupied sequence slots). Step spacing reflects elapsed time. Dotted lines mark step boundaries. Training time, credit-assignment latency, staleness expiry, and replay are excluded.}
    \label{fig:mixrl_dispatch}
\end{figure}

\paragraph{Adaptive Rollout Scheduling}
Within these budgets, scheduling balances long-run generation demand with progress toward the current training batch. 
If $A_i$ groups have already been accepted from source $i$ for the current batch, its scheduling weight is
\begin{equation}
    w_i=\alpha\frac{B_i}{r_i}
        +(1-\alpha)\frac{(B_i-A_i)^+}{r_i},
    \label{eq:mixrl_swrr}
\end{equation}
where $(x)^+=\max(x,0)$ and $\alpha\in[0,1]$. 
The target term maintains generation demand; the deficit term prioritizes sources with a remaining deficit.
These weights drive smooth weighted round-robin; \textit{Steady-state startup} combines $\alpha=0.5$ with initial concurrency allocated in proportion to $t_i m_i$.
The trace-driven simulation in Figure~\ref{fig:mixrl_dispatch} compares the policies at a fixed concurrency limit with nonbinding source budgets: \textit{Deficit-corrected} scheduling ($\alpha=0.5$) improves occupancy stability over \textit{Deficit-based} scheduling ($\alpha=0$) and collection balance over \textit{Target-based} scheduling ($\alpha=1$) in this workload.
Simulation durations are calibrated to each source's mean rollout time at \textit{End} in Figure~\ref{fig:mixrl_heterogeneity}.

\paragraph{Predictive Rollout Dispatch}
We jointly estimate KV demand and expected inference concurrency to guide the admission and placement of new rollouts across ranks. 
Per-source priors estimate a rollout's total sequence length (input and generated tokens), summed across contexts; these priors are updated from completed rollouts and used to estimate KV demand.
We admit a rollout only when a safety-margin multiple of its estimated KV demand fits within the target rank's remaining GPU KV capacity.
Hierarchical caching (\S\ref{sec:rlinfra:traininf}) spans HBM and a pinned host pool: the two tiers jointly retain the state of admitted rollouts.
The estimated fraction of rollout time spent in environment execution converts rollout concurrency into expected inference concurrency.
We bound this expectation by the max running requests used for CUDA graph capture, limiting queueing delays that prolong rollout lifetimes and increase staleness.
To maximize throughput under both constraints, a greedy heuristic selects the feasible rank with the greatest remaining capacity---the minimum of available concurrency slots and remaining KV capacity, both expressed in sequence units---with both quantities updated after each placement.

\paragraph{Sample Replay}
We use sample replay to accelerate initial collection from slow sources. 
In profiling, startup sample collection took approximately $1.8\times$ as long as that of continuous operation.
Within slow sources, shorter rollouts tend to finish first, biasing the initial batch even when per-source targets are met.
We therefore reuse completed rollout groups from selected slow sources during the first collection step after startup or checkpoint recovery, filling remaining per-source deficits under current filtering rules.
Fresh-start replay assumes that stored rollouts were generated under the run's starting policy; recovery replay reuses completed rollouts generated before recovery and still within each source's \textit{staleness} limit.
Restricting replay to the first collection step reduces the wait for slow sources while preserving the specified training distribution.

\subsection{Training/Inference Consistency and Optimization}
\label{sec:rlinfra:traininf}
We extend the RL and OPD infrastructure of MiMo-V2-Flash~\citep{mimo2026v2flash}, keeping SGLang~\citep{zheng2024sglang} and Megatron-LM~\citep{shoeybi2019megatron} as the inference and the training engine, respectively.
In \mimox{}, we use MXFP4 as the experts' data type during rollout.
We maintain alignment between the two engines in mixture-of-experts (MoE) routing and probability normalization.
On the inference side, the \textit{Context Cache} carries routing records, sampling candidate sets, and visual inputs alongside the KV state across turns, and offloads idle state to host memory.
A draft model trained on RL rollout logs further accelerates generation through speculative decoding.
On the training side, we reduce the memory and communication costs of long sequences.

\paragraph{Training--Inference Consistency}
To support the \mimox{} series, we apply quantize--dequantize (QDQ) to the experts after each parameter update.
The quantization follows the numeric constraints of the MXFP4 Humming GEMM kernels~\citep{vllm_humming_2026} used during rollout, so the two engines see identical expert weights.
Even under identical parameters, numerical differences between the engines can flip discrete expert selections; Rollout Routing Replay (R3)~\citep{ma2025stabilizing} records the expert indices used during rollout and replays them during training, reproducing the captured execution path.
Top-\textit{k} and top-\textit{p} sampling renormalize over a restricted candidate set rather than the full vocabulary; we record each token's candidate set during rollout~\citep{liu2025deepseek,mai_thinking_1} and renormalize the training log-probabilities within it, matching the sampler's normalization.
For top-\textit{p} sampling, only the GPU--CPU transfer is dense: we ship a bitmap of fixed shape and full-vocabulary width. 
The fixed shape avoids a GPU--CPU synchronization; the full width never truncates even a set that spans the whole vocabulary.
All later stages are sparse: at a typical top-\textit{p} of 0.97, a candidate set averages fewer than five tokens.
Both R3 and top-\textit{p} candidate-set replay incur negligible overhead, as the payloads stay compact and move off the critical path.

\paragraph{Context Caching}
Following the request-level KVCache of MiMo-V2-Flash~\citep{mimo2026v2flash}, each dialogue context has a persistent key: within one policy version, later turns hit the cached KV---including generated tokens---and prefill only the new suffix.
Context caching is a deliberately stateful choice: the cached state outlives each turn.
Expert indices and candidate sets are therefore not returned on intermediate turns but only when the rollout is finally collected, avoiding fragmented per-turn communication and processing.
Historical multimodal inputs need neither re-hashing nor re-sending: their visual tokens already sit inside the cached KV.
Only newly introduced images cross the process boundary; full visual inputs are sent only after a policy update or cache miss.
Multi-turn environment interaction divides a trajectory's wall clock into GPU time (generation) and tool time (waiting on the environment); a hierarchical extension of the cache reflects this temporal split \textit{spatially}---a state resides in HBM during GPU time and in a pinned host pool during tool time.
Offloading and restoration run on side CUDA streams rather than the compute stream, never stalling generation.
HBM is thereby devoted to running requests as far as possible: the larger the decode batch, the higher the arithmetic intensity and compute utilization.

\paragraph{Draft Model Acceleration}
RL rollouts use block-$6$ DFlash for speculative decoding by default, replacing the multi-token prediction (MTP-$3$) configuration inherited from SFT.
Initially trained on the SFT policy, DFlash is then finetuned on early RL rollout logs resampled to match the RL training distribution.
Under this default configuration, average accepted length is 31.3\% higher than with the MTP configuration.
At large RL batch sizes, we select the draft block size based on end-to-end throughput rather than acceptance rate alone.
Smaller draft blocks reduce verification work: in mixed-task RL evaluation, block-$6$ improves global average throughput by approximately 6\% over block-$8$, with little change in average accepted length.
Low-precision draft computation (FP8) is also applied to reduce drafting overhead.
On our long-context workload, RL-adapted FP8 DFlash achieves approximately 10.3\% higher per-node throughput than the baseline.

\paragraph{Training Optimization} 
We train RL at a 1M-token context length.
\mimox{} interleaves 128-token sliding-window layers with full attention.
Under context parallelism, the sliding-window layers exchange only the KV their queries can reach---at most a window-sized segment---so per-layer traffic is bounded by the window rather than the sequence length.
Long sequences also inflate the memory footprint, and MoE expert imbalance pushes peak memory higher.
Optimizer states stay in CPU memory and are copied back only for parameter updates.
All loss computation is fused into one kernel: the policy-gradient loss (with or without top-\textit{p} renormalization) and the OPD loss, optionally with metrics such as entropy, label logit, and top-\textit{p} mass.
The fusion saves memory and reduces step time.

%% file: 8-open.tex
\providecolor{xiaomiorange}{HTML}{FF6901}

\section{Open Foundations for Agentic RL}
\label{sec:open}

The MiMo-V2.6 series demonstrates the potential of reinforcement learning to substantially advance agentic capabilities. Building on this progress requires capable models and high-quality RL environments that pair meaningful tasks with reliable verifiers. Sharing these resources alongside reproducible baselines supports continued research and innovation in the open-source community. 
We therefore open-source MiMo-V2.6-Distill-Qwen-9B,\footnote{\url{https://huggingface.co/XiaomiMiMo/MiMo-V2.6-Distill-Qwen-9B}} a small model distilled from MiMo as a shared starting point for further RL training. Alongside the model, we release high-quality RL environments and an end-to-end open-source RL framework. In this section, we establish domain-specific GRPO baselines using these released resources. We also conduct a separate multi-harness training experiment using coding as a case study, following the approach described for MiMo-V2.6. Our experiments show substantial gains across multiple domains, demonstrating the value of these resources for RL training.

\subsection{Distillation from MiMo-V2.6}
\label{sec:mimo-distilled-qwen-9b}

Scaling RL computation and broadening the coverage of environments
and harnesses create more opportunities for exploration and learning.
Across these settings, effective exploration depends on coordinating
actions and responding to environmental feedback. Realizing the
potential of agentic RL therefore begins with a strong foundation
of task-solving capabilities.

To provide the open-source community with such a foundation,
we develop MiMo-V2.6-Distill-Qwen-9B by transferring MiMo's agentic
experience into a smaller model. We obtain the model by supervised
fine-tuning Qwen3.5-9B~\citep{qwen3next2025} on MiMo-generated data spanning coding, general-domain, visual, and cybersecurity
tasks. The mixture contains 77.4 billion total tokens, including
27.2 billion loss tokens that contribute to the SFT objective.
Table~\ref{tab:mimo-sft-data} summarizes the data composition.

\begin{table}[t]
    \centering
    \small
    \setlength{\tabcolsep}{4.2pt}
    \renewcommand{\arraystretch}{1.12}
    \renewcommand{\tabularxcolumn}[1]{m{#1}}
    \begin{tabularx}{\linewidth}{@{}>{\raggedright\arraybackslash}X *{3}{>{\centering\arraybackslash}X}@{}}
        \toprule
        \textbf{Data source}
                    & \makecell{\textbf{Total tokens}\\\textbf{(B)}}
                    & \makecell{\textbf{Token share}\\\textbf{(\%)}}
                    & \makecell{\textbf{Loss tokens}\\\textbf{(B)}} \\
        \midrule
        Code    & 23.2 & 29.9 & 7.3 \\
        Cyber   & 11.0 & 14.2 & 4.8 \\
        General & 22.0 & 28.5 & 5.7 \\
        Visual  & 21.2 & 27.4 & 9.4 \\
        \midrule
        \textbf{Total} & \textbf{77.4} & \textbf{100.0} & \textbf{27.2} \\
        \bottomrule
    \end{tabularx}
    \caption{Weighted SFT data composition. Token counts are in billions.
    Values are rounded to one decimal place; totals use unrounded
    aggregate counts.}
    \label{tab:mimo-sft-data}
\end{table}

The resulting MiMo-V2.6-Distill-Qwen-9B checkpoint improves over
Qwen3.5-9B across the reported evaluations
(Table~\ref{tab:open-rl-results}). For example, SWE-bench Pro~\citep{deng2025swe} increases from
32.0 to 44.6, and AutomationBench from 5.0\% to 30.3\%.
We view these strengthened task-solving capabilities as a
promising foundation for further improvement through RL.
We therefore use this checkpoint as the common initialization
for the domain-specific GRPO experiments and the multi-harness
coding experiment described below.

\subsection{RL with Open Environments}
\label{sec:open-rl}

\begin{table}[!t]
    \centering
    \small
    \setlength{\tabcolsep}{4.2pt}
    \renewcommand{\arraystretch}{1.12}
    \begin{tabular*}{\linewidth}{@{\extracolsep{\fill}}l l c l@{}}
        \toprule
        \textbf{Domain}
          & \textbf{Task Family}
          & \textbf{Training Tasks}
          & \textbf{Verifier} \\
        \midrule
        Code    & Software engineering & 3k & Executable tests \\
        Cyber   & Vulnerability reproduction & 1k & Rule checks \\
        General & Knowledge work & 1k & Rubric-based judging \\
        Visual  & Web development & 2k & Visual grading \\
        \bottomrule
    \end{tabular*}
    \caption{Overview of the released RL environments, including training
    task counts and the verifiers used to evaluate task completion.
    Task counts are approximate, based on distinct task identifiers in
    each training set; k denotes one thousand.}
    \label{tab:open-rl-environments}
\end{table}

\paragraph{Environments Overview}
\label{sec:open-environments}
Table~\ref{tab:open-rl-environments} summarizes the RL environments we
release across coding, cybersecurity, general-domain, and visual tasks.
The four training sets contain approximately 7k tasks in total,
with the general-domain set focusing on knowledge work.
The training resources additionally include approximately 1k
music-generation tasks, which support the GRPO experiments reported below.

\paragraph{RL Experiments}
\label{sec:open-rl-experiments}
Starting from the same MiMo-V2.6-Distill-Qwen-9B SFT checkpoint, we
conduct GRPO training separately for coding, cybersecurity,
general-domain, and visual tasks using the corresponding released
environments. 
Alongside public benchmarks, we evaluate on four internal sets:
MiMo Code Bench (mini) for software engineering , MiMo Cyber Bench (mini)
for vulnerability reproduction, MiMo General Bench (mini) for knowledge work, and MiMo Visual Coding (mini) for website development. These
evaluation sets follow the same task distributions as their corresponding training sets.

\begin{table}[t]
    \centering
    \small
    \setlength{\tabcolsep}{4.2pt}
    \renewcommand{\arraystretch}{1.12}
    \renewcommand{\tabularxcolumn}[1]{m{#1}}
    \begin{tabularx}{\linewidth}{@{}l c | >{\centering\arraybackslash}X | *{2}{>{\centering\arraybackslash}X}@{}}
        \toprule
        \multirow{2}{*}{\textbf{Benchmark}}
          & \multirow{2}{*}{\textbf{Metric}}
          & \multicolumn{1}{c|}{\multirow{2}{*}{\textbf{Qwen3.5-9B}}}
          & \multicolumn{2}{c}{\textbf{MiMo-V2.6-Distill-Qwen-9B}} \\
        \cmidrule(lr){4-5}
          & & \multicolumn{1}{c|}{}
            & \multicolumn{1}{c}{\textbf{SFT}} & \textbf{RL} \\
        \midrule
       \rowcolor{xiaomiorange!10}[0pt][0pt]
        \multicolumn{5}{@{}l@{}}{\textbf{Code}} \\
        SWE-bench Verified      & avg@3      & 60.0 & 61.1 & \textbf{66.2} \\
        SWE-bench Pro           & avg@3      & 32.0 & 44.6 & \textbf{47.6} \\
        MiMo Code Bench (mini)      & avg@3      & 19.5 & 51.6 & \textbf{59.9} \\
        \midrule
       \rowcolor{xiaomiorange!10}[0pt][0pt]
        \multicolumn{5}{@{}l@{}}{\textbf{Cyber}} \\
        MiMo Cyber Bench (mini)        & avg@3      &  5.7 & 31.3 & \textbf{47.0} \\
        \midrule
       \rowcolor{xiaomiorange!10}[0pt][0pt]
        \multicolumn{5}{@{}l@{}}{\textbf{General}} \\
        AutomationBench v1.0.6   & avg@1      &  5.0 & 30.3 & \textbf{33.1} \\
        Terminal Bench 2.1 & avg@1      & 27.0 & 37.1 & \textbf{52.8} \\
        Toolathlon-Verified          & avg@1      & 25.9 & 35.2 & \textbf{38.0} \\
        OfficeQA Pro          & avg@1      &  9.0 & 19.5 & \textbf{24.8} \\
        JobBench          & avg@1      &  2.6 & 18.3 & \textbf{25.2} \\
        MiMo General Bench (mini)      & avg@1      & 28.5 & 62.2 & \textbf{70.6} \\
        \midrule
        \rowcolor{xiaomiorange!10}[0pt][0pt]
        \multicolumn{5}{@{}l@{}}{\textbf{Visual}} \\
        MiMo Visual Coding (mini)       & avg@1      & 61.7 & 64.0 & \textbf{72.4} \\
        \bottomrule
    \end{tabularx}
    \caption{Evaluation of Qwen3.5-9B, MiMo-V2.6-Distill-Qwen-9B after SFT,
    and the domain-specific checkpoints after further GRPO training (RL).
    Code uses single-harness RL. The highest available
    score in each row is \textbf{bolded}.}
    \label{tab:open-rl-results}
\end{table}


Table~\ref{tab:open-rl-results} shows that RL improves on the SFT
checkpoint in all 11 evaluations reported in the table, spanning the
four task domains. SWE-bench Verified~\citep{jimenez2024swebench} rises from 61.1 to 66.2, while
Terminal~Bench~2.1~\citep{merrill2026terminalbench} improves from 37.1 to 52.8. OfficeQA Pro~\citep{officeqa} improves from 19.5 to 24.8, and Toolathlon-Verified~\citep{hkustnlp2026toolathlonverified} increases from 35.2 to 38.0. MiMo Visual Coding (mini) improves from
64.0 to 72.4, and
MiMo Cyber Bench (mini) increases from 31.3 to 47.0. Beyond the tasks reported in Table~\ref{tab:open-rl-results}, we also explore emerging tasks in artistic creation and design, such as music composition. On our internal music benchmark, the score improves substantially from 45.7 after SFT to 52.5 after RL. These results highlight the value
of high-quality training data and a strong SFT initialization for
further improvement through RL.

\paragraph{Multi-Harness Training}
\label{sec:open-multi-harness}
The coding results in Table~\ref{tab:open-rl-results} are obtained through
single-harness RL. We further investigate multi-harness training in a
separate coding experiment, starting from the same
MiMo-V2.6-Distill-Qwen-9B SFT checkpoint. Following the
multi-harness training approach described for MiMo-V2.6,
we jointly optimize the model across four mini-harnesses and evaluate
it on these harnesses and three additional held-out harnesses.
Table~\ref{tab:open-multi-harness} compares Qwen3.5-9B with
MiMo-V2.6-Distill-Qwen-9B before and after multi-harness RL across three
coding evaluations and seven agent harnesses.
MiMo-V2.6-Distill-Qwen-9B improves on Qwen3.5-9B in all 21 dataset--harness
pairs, and multi-harness RL further improves every pair. On MiMo Code Bench (mini), the additional gains over SFT range from 1.8 to 9.3 percentage
points across the seven harnesses.

\begin{table}[!t]
    \centering
    \small
    \setlength{\tabcolsep}{3.8pt}
    \renewcommand{\arraystretch}{1.12}
    \resizebox{\linewidth}{!}{%
    \begin{tabular}{@{}l *{8}{>{\centering\arraybackslash}m{4.1em}}@{}}
        \toprule
        & \multicolumn{4}{c}{\textbf{Training harnesses}}
        & \multicolumn{3}{c}{\textbf{Held-out harnesses}}
        & \\
        \cmidrule(lr){2-5}\cmidrule(lr){6-8}
        \textbf{Model}
          & {\footnotesize\makecell{\textbf{mini-}\\\textbf{harness1}}}
          & {\footnotesize\makecell{\textbf{mini-}\\\textbf{harness2}}}
          & {\footnotesize\makecell{\textbf{mini-}\\\textbf{harness3}}}
          & {\footnotesize\makecell{\textbf{mini-}\\\textbf{harness4}}}
          & {\footnotesize\textbf{codex}}
          & {\footnotesize\makecell{\textbf{claude}\\\textbf{code}}}
          & {\footnotesize\makecell{\textbf{mini-swe-}\\\textbf{agent}}}
          & {\footnotesize\textbf{Mean}} \\
        \midrule
       \rowcolor{xiaomiorange!10}[0pt][0pt]
        \multicolumn{9}{@{}l@{}}{\textbf{SWE-bench Verified}} \\
        Qwen3.5-9B
          & 58.4 & 36.4 & 54.8 & 57.8
          & 48.6 & 54.8 & 60.6 & 53.1 \\
        MiMo-V2.6-Distill-Qwen-9B
          & 61.7 & 63.9 & 61.7 & 63.1
          & 58.5 & 61.7 & 65.3 & 62.3 \\
        \quad + Multi-Harness RL
          & \textbf{67.9} & \textbf{65.1} & \textbf{66.6} & \textbf{67.2}
          & \textbf{61.1} & \textbf{65.3} & \textbf{66.7} & \textbf{65.7} \\
        \midrule
       \rowcolor{xiaomiorange!10}[0pt][0pt]
        \multicolumn{9}{@{}l@{}}{\textbf{SWE-bench Pro}} \\
        Qwen3.5-9B
          & 33.5 & 15.2 & 31.1 & 31.1
          & 23.1 & 26.9 & 31.6 & 27.5 \\
        MiMo-V2.6-Distill-Qwen-9B
          & 45.2 & 46.6 & 45.1 & 45.5
          & 40.3 & 42.2 & 45.6 & 44.4 \\
        \quad + Multi-Harness RL
          & \textbf{48.5} & \textbf{46.9} & \textbf{47.6} & \textbf{48.0}
          & \textbf{42.6} & \textbf{43.1} & \textbf{48.4} & \textbf{46.5} \\
        \midrule
        \rowcolor{xiaomiorange!10}[0pt][0pt]
        \multicolumn{9}{@{}l@{}}{\textbf{MiMo Code Bench (mini)}} \\
        Qwen3.5-9B
          & 19.0 & 7.5 & 16.5 & 22.5
          & 10.5 & 13.5 & 17.5 & 15.3 \\
        MiMo-V2.6-Distill-Qwen-9B
          & 53.2 & 56.7 & 51.5 & 56.3
          & 46.0 & 51.2 & 56.8 & 53.1 \\
        \quad + Multi-Harness RL
          & \textbf{62.5} & \textbf{64.5} & \textbf{57.0} & \textbf{63.3}
          & \textbf{50.7} & \textbf{53.0} & \textbf{62.0} & \textbf{59.0} \\
        \bottomrule
    \end{tabular}}
    \caption{Coding performance across agent harnesses.
    Mean is the unweighted average of the seven harness scores, rounded
    to one decimal place. The best result in each column within a
    dataset is \textbf{bolded}.}
    \label{tab:open-multi-harness}
\end{table}

\begin{figure}[t]
    \centering
    \includegraphics[width=0.98\linewidth]{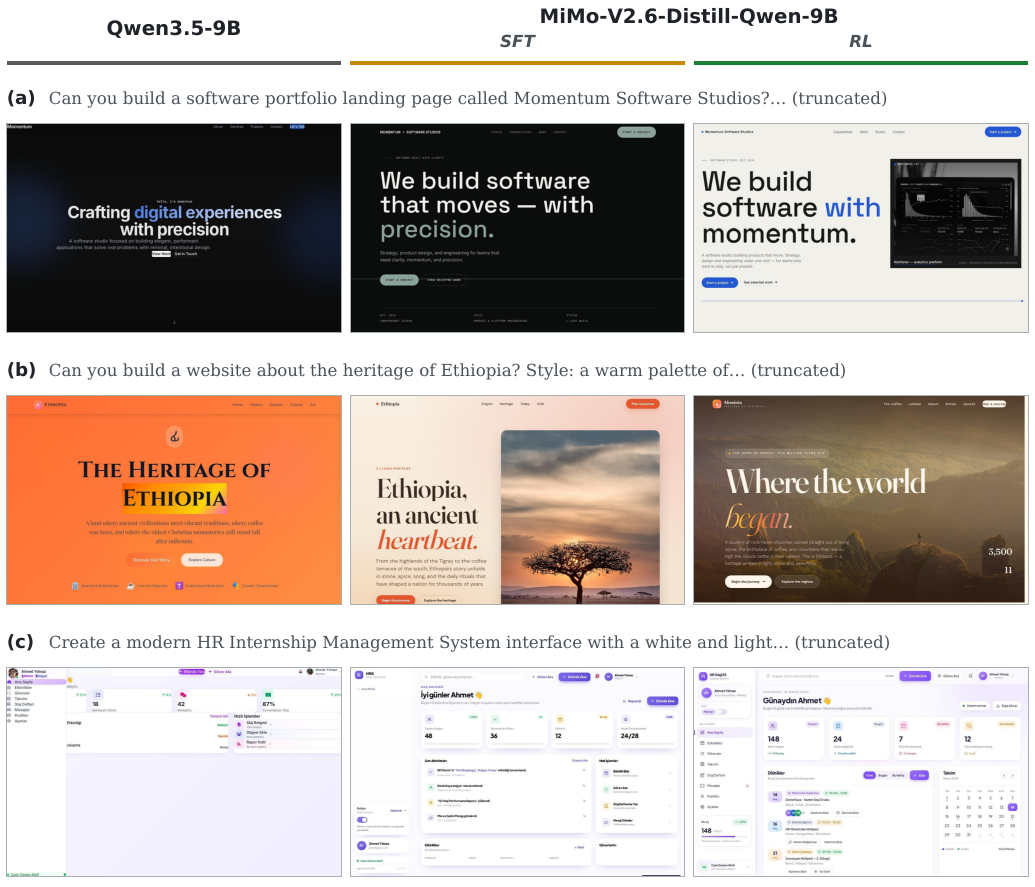}
    \caption{Examples of website hero sections generated by Qwen3.5-9B, MiMo-V2.6-Distill-Qwen-9B (SFT), and MiMo-V2.6-Distill-Qwen-9B (RL). Each row (a–c) shows outputs from the three models for the same prompt.}
    \label{fig:qwen_webdev_showcase}
\end{figure}

\paragraph{Case Study}
\label{sec:open-case-study}
To offer a more intuitive view of how our SFT and RL recipe progressively enhances model capabilities, we present a case study on web development (a domain where visual quality is immediately apparent), comparing websites generated by Qwen3.5-9B, MiMo-V2.6-Distill-Qwen-9B (SFT), and MiMo-V2.6-Distill-Qwen-9B (RL) in Figure~\ref{fig:qwen_webdev_showcase}. As shown in the left column, websites produced by Qwen3.5-9B appear relatively plain in visual design, with simplistic layouts and minimal use of imagery. Notably, the HR management interface in (c) suffers from clear layout issues. After distillation, MiMo-V2.6-Distill-Qwen-9B (SFT) (middle column) generates websites with noticeably richer content, more harmonious color palettes, and more effective incorporation of image assets, as exemplified by the photographic elements in the Ethiopia heritage page in (b). Following RL training (right column), the model takes a further step, delivering the most polished results with refined typography, visually striking hero sections (e.g., the sunset imagery in~(b)), and more complete, well-structured page layouts (e.g., the fully featured HR dashboard in~(c) with sidebar navigation, a calendar widget, and detailed statistics cards). These progressive qualitative improvements align with the quantitative
trajectory ($61.7 \rightarrow 64.0 \rightarrow 72.4$) observed in Table~\ref{tab:open-rl-results}, jointly demonstrating the cumulative benefits of distillation from MiMo-V2.6 followed by RL in enhancing both the aesthetic quality and functional completeness of generated outputs.

%% file: author.tex
\section{Contributions and Acknowledgments}
We would like to express our sincere gratitude to all contributors for their invaluable support and
efforts, including the Xiaomi Data Platform, CloudML, NGK, MiChat, Mify, MiKS and LLM-Plus teams, as
well as those not explicitly listed in this paper.
\textit{Within each role, authors are listed in reverse alphabetical order by first name.}

{\renewcommand{\thefootnote}{\fnsymbol{footnote}}\footnotetext[2]{Corresponding author}}

\textbf{Core Contributors:} Zongming Qiao, Ziyue Hua, Zirui Ou, Zihao Yue, Zihan Jiang, Zhuo Huang, Zhiyang Chen, Zhixian Zheng, Zhipeng Xu, Zhengrui Ma, Yuyang Hu, Yuhang Dong, Yuechen Zhang, Yudong Wang, Yuanxin Liu, Yixin Yang, Yishuo Cai, Yikai Zhao, Yihan Yan, Yifan Zhang, Yifan Song, Xiyu Wei, Xing Zhang, Xin Zhang, Xiaoqian Liu, Xiaodong Ji, Xiangwei Deng, Xueyu Guo, Wenhan Ma, Weimin Xiong, Weikun Wang, Weiji Zhuang, Shuo Liu, Shuhuai Ren, Shuhao Gu, Shimao Chen, Shijie Cao, Shihua Yu, Shicheng Li, Shengjie Zhou, Shaolei Zhang, Rang Li, Qiying Wang, Qingkai Fang, Qianli Chen, Minzheng Wang, Liwen Wang, Linli Yao, Linghao Zhang, Liangyu Cheng, Liang Zhao, Lei Li, Jinhao Dong, Jinyu Xiang, Jianyu Wei, Jiangshan Duo, Huaqiu Liu, Huanjie Fan, Hongyi Guan, Hongshen Xu, Hao Tian, Hanyu Li, Hailin Zhang, Gang Wang, Fuli Luo\textsuperscript{$\dagger$}, Feng Wei, Dong Zhang, Dawei Zhu, Chiheng Lou, Chenhong He, Chenhao He, Chenghua Liu, Bowen Ye, Bowen Shen, Boshen Xu, Bo Yang, Bingquan Xia, Bangjun Xiao, Baixuan Xu

\textbf{Contributors:} Zhouxiang Mao, Zhiyang Zhang, Zhixiang Xu, Zhenru Lin, Zhengju Tang, Zhaojun Huang, Yuzhe Weng, Yuxing Xiang, Yuxiao Li, Yuheng Yang, Yuhang Wang, Yuchen Liu, Yuanyuan Tian, Yuanliang Dong, Yu Cheng, Yongzhe He, Yongshun Liang, Yong Wang, Yiyan Wang, Yitian Gong, Yijie Zhang, Yanshu Xin, Xun Zhang, Xingjian Zhao, Wenyu Yang, Wenshan Huang, Wenhao Li, Tingwei Huang, Tianyu Yu, Tianyang Lu, Taoyu Yang, Sinan Du, Shutong Tian, Shulin Du, Shengfan Wang, Shanchuan Fang, Qihao Zhang, Qibin Yang, Qian Yu, Qian Tu, Pengrong Xie, Peipei Wang, Peidian Li, Minkun Guo, Mingchen Shao, Luohan Gao, Lijie Wang, Liang Shi, Kaiqi Chen, Kaiming Liu, Kaifei Wang, Kai Yang, Jinlong Xue, Jiechen Zhang, Jiaxuan Liu, Hongxu An, Hao Peng, Hanglong Lü, Guonan Wang, Feiyu Yang, Fanyu Cao, Fangyue Liu, Fan Cui, Cong Wang, Chun Chen, Chenxu Bai, Chengxuan Zhu, Chenghua Wang, Boyi Zeng
